\documentclass[journal,10pt,twoside, a4paper]{IEEEtran} 

\IEEEoverridecommandlockouts                              
\usepackage{amsmath}      
\usepackage{graphicx}
\usepackage{subcaption}  
\usepackage{float}  
\usepackage[dvipsnames]{xcolor}
\usepackage{booktabs} 
\usepackage{amsmath, amssymb, algorithm, algpseudocode, caption}

\usepackage{amsmath}
\usepackage{amssymb}
 \usepackage{multirow}

\begin{document}

\title{
ON-Traffic: An Operator Learning Framework for Online Traffic Flow Estimation and Uncertainty Quantification from Lagrangian Sensors
}

\author{Jake Rap, Amritam Das
\thanks{Eindhoven University of Technology, Department of Electrical Engineering, Control Systems group, P.O.~Box 513, 5600 MB Eindhoven, The Netherlands. E-mail: {\tt\small j.e.w.rap@tue.nl, am.das@tue.nl}

$^*$Corresponding author: Amritam Das.}
}

\maketitle
\thispagestyle{empty}
\pagestyle{empty}

\begin{abstract}
Accurate traffic flow estimation and prediction are critical for the efficient management of transportation systems, particularly under increasing urbanization. Traditional methods relying on static sensors often suffer from limited spatial coverage, while probe vehicles provide richer, albeit sparse and irregular data. This work introduces ON-Traffic, a novel deep operator Network and a receding horizon learning-based framework tailored for online estimation of spatio-temporal traffic state along with quantified uncertainty by using measurements from moving probe vehicles and downstream boundary inputs. Our framework is evaluated in both numerical and simulation datasets, showcasing its ability to handle irregular, sparse input data, adapt to time-shifted scenarios, and provide well-calibrated uncertainty estimates. The results demonstrate that the model captures complex traffic phenomena, including shockwaves and congestion propagation, while maintaining robustness to noise and sensor dropout. These advancements present a significant step toward online, adaptive traffic management systems.

\end{abstract}

\section{INTRODUCTION}

    
Estimation and prediction of traffic flow is essential for the management of transportation systems, as urbanization accelerates and mobility demands grow. Reliable traffic prediction enables the implementation of predictive control and decision-making strategies to mitigate congestion, reduce pollution, and shorten travel times, as well as provide valuable insight for analyzing traffic patterns and trends. 

Traditional traffic flow estimation often relies on fixed infrastructure sensors, such as loop detectors, cameras, and radar systems, which are strategically placed at fixed points in the road network. Although these methods provide valuable data on traffic variables such as density, speed, and flow, they are inherently limited in spatial coverage because they can monitor only the areas where sensors are installed. Although there is a growing number of such sensors in use, full coverage of entire road networks remains a challenge. 

To address these limitations, recent studies \cite{barreau2020learningbasedstatereconstructionscalar}, \cite{cicic2020numericalinvestigationtrafficstate} have explored the incorporation of probe vehicles into traffic data collection systems. Probe vehicles, equipped with sensors to detect their position, velocity, and intervehicular distance, offer a mobile and dynamic means of capturing traffic data that complements the static data from traditional sensors. This integration enables a more comprehensive view of the traffic network, combining the fixed nature of infrastructure sensors with the flexibility and coverage of moving vehicles. 

\subsection{Main Challenges in Traffic Flow Estimation}
\subsubsection{Data-type} Despite the fact that introducing probe vehicles on top of static sensors offers more information about traffic, it also presents challenges, as their data is often sparse, irregular and has a high degree of variability. These characteristics pose a significant hurdle for using the data in both traditional and learning-based algorithms for estimating the traffic-state.

\subsubsection{Online surrogation} Moreover, in real-world applications, it is crucial to perform traffic flow estimation and prediction in an online manner. In this context, "online" refers to the continuous processing of data as it becomes available, allowing the systems to update traffic predictions dynamically without the need for large-scale computations or retraining. 

\subsubsection{Quantifying aleatoric uncertainty} In addition, it is also essential to quantify the inherent aleatoric uncertainties in the probe vehicle data, which are typically incomplete and noisy, and can make the problem of traffic flow prediction ill-posed. Therefore, Uncertainty Quantification (UQ) is key to a reliable prediction of traffic density with confidence estimates and to identifying areas where the traffic predictions are less trustworthy.

These combined challenges call for the need of traffic estimation approaches that are not only robust to sparse and irregular data, but also capable of online deployment and equipped with methods for quantifying uncertainty.


\subsection{State-of-the-Art}

Traffic flow estimation has been extensively studied using mathematical models, with the Lighthill-Whitham-Richards (LWR) model \cite{lighthill1955}, \cite{richards1956}, being a foundational first principle model in this field. This Partial Differential Equation (PDE), based on the conservation of vehicles and a fundamental diagram relating traffic density to velocity, describes the spatio-temporal model exhibiting key traffic phenomena such as shockwaves and congestion formation.

Numerical methods, such as the Godunov scheme \cite{Friedrich_2018}, have traditionally been employed to solve the LWR model for the forward problem, which involves predicting traffic flow from known initial conditions. These methods are highly effective under structured data conditions, utilizing discretized spatial and temporal grids. However, their reliance on regular grids limits their applicability in real-world traffic monitoring, where data from probe vehicles are sparse, irregular, and mobile. For the inverse problem, reconstructing traffic flow from incomplete, trajectory-based data, these numerical methods face significant challenges and, therefore, cannot be applied directly to such dynamic and irregular inputs we consider \cite{particlefilter}.

Learning-based approaches, particularly those utilizing artificial neural networks, offer greater flexibility compared to traditional methods. Several studies have applied Physics-Informed Neural Networks (PINNs) to leverage sparse and irregular data while incorporating traffic physics. PINNs utilize governing equations, such as the conservation laws in the LWR model, to reconstruct traffic flow from incomplete data. For instance, \cite{pmlr-v144-barreau21a} and \cite{liu2020learning} employed PINNs with simulated probe vehicle data from the Simulation of Urban MObility (SUMO) simulator \cite{SUMO2018}, while \cite{shi2020ARZ} extended this approach to the second-order Aw-Rascle-Zhang (ARZ) \cite{aw2000resurrection} model, incorporating the conservation of both mass and momentum for greater accuracy. Additionally, \cite{barreau2021physicsinformedlearningidentificationstate} enhanced PINNs by learning the fundamental diagram of traffic flow and introducing physics-based loss terms to suppress biased noise during training. Despite these advancements, traditional PINNs are fundamentally limited to the specific scenarios or initial conditions on which they are trained. Retraining is required to estimate traffic for new scenarios, making conventional PINNs unsuitable for online prediction.

To overcome this limitation and enable accurate traffic flow estimation in dynamic conditions, we turn to operator learning as a potential solution. Operator learning methods, such as DeepONets \cite{osti_2281727}, have been developed to map functions to functions, enabling generalization across different scenarios by learning the underlying spatio-temporal behavior of the dynamical system. Although operator learning shows significant potential, particularly for developing online surrogate models that generalize across multiple scenarios, its application to traffic flow estimation is still in its early stages. Notable successes in other domains, such as gas flow dynamics \cite{Kumar2024} and ship motion prediction \cite{shorttermship}, demonstrate its promise. However, directly applying standard operator learning methods to traffic flow faces challenges due to the variability and irregularity of real-world data.

Recent advancements in operator learning provide numerous upgrades over the classical DeepONet architecture. One such advancement is the Variable-Input Deep Operator Network (VIDON) \cite{prasthofer2022variableinputdeepoperatornetworks}, which is designed to handle irregular input data, such as varying numbers and locations of sensors. VIDON employs an attention mechanism inspired by transformers \cite{transformer} to dynamically process sensor values and their positions. Attention weights are computed between encoded inputs, capturing relationships between sensors in a permutation-invariant manner. This flexibility allows VIDON to adapt to dynamic sensor configurations, making it particularly suitable for irregular data scenarios. Similar approaches, such as LOCA \cite{loca} and GraphDeepONet \cite{graphdeeponet}, incorporate graph neural networks and attention mechanisms, but they often lack the adaptability of VIDON in handling highly variable input sizes and locations. As a result, VIDON has the potential to serve as the baseline for our proposed extensions to develop operator learning for traffic flow estimation.

On the other hand, quantifying aleatoric uncertainty is well studied in deep learning. For example, Bayesian Neural Networks (BNNs) \cite{magnani2022approximatebayesianneuraloperators} offer a probabilistic interpretation of outputs by modeling neural network weights as probability distributions. However, their computational complexity and scalability related issues make them less practical for operator learning tasks. Conv-PDE-UQ \cite{convpdeUQ} and Probabilistic DeepONet \cite{deeponetUQ} address these challenges by drawing inspiration from Gaussian Processes \cite{GP}. These methods extend the standard operator formulation by introducing an uncertainty operator to predict both outputs and their associated variances. This dual prediction framework not only provides reliable estimates but also enhances the applicability of operator learning in scenarios where understanding uncertainty is critical.

\subsection{Main Contribution}
The objective of this research is to resolve the lingering difficulties in online estimation of spatio-temporal
traffic state with quantified measure of reliability while using only sparse, irregular and variable measurements from moving probe vehicles. To this end, we introduce ON-Traffic, an online surrogate operator that resolves these difficulties and predicts the traffic flow density and velocity profile along with quantified aleatoric uncertainty bound. With respect to the state-of-the-art learning frameworks, our proposed methodology is positioned as in Tab. \ref{tab:network_tasks}.

\begin{table}[H]
    \centering
    \caption{Comparison of Network Architectures related to the challenges we tackle.}
    \renewcommand{\arraystretch}{1.2} 
    \normalsize
    \scalebox{0.95}{
        \begin{tabular}{l c c c}
            \toprule
            \textbf{Network} & \textbf{Irregular} & \textbf{Uncertainty} & \textbf{Online} \\
            \textbf{Architecture} & \textbf{Data} & \textbf{Quantification} & \textbf{Surrogation} \\
            \midrule
            PINN & \checkmark & \checkmark & \texttimes \\
            DeepONet & \texttimes & \texttimes & \checkmark \\
            Prob-DeepONet & \texttimes & \checkmark & \checkmark \\
            VIDON & \checkmark & \texttimes & \checkmark \\
            \textcolor{teal}{ON-Traffic} & \textcolor{teal}{\checkmark} & \textcolor{teal}{\checkmark} & \textcolor{teal}{\checkmark} \\
            \bottomrule
        \end{tabular}
        }
    \label{tab:network_tasks}
\end{table}

Specifically, the innovative features of ON-Traffic are as follows:

\begin{enumerate}
    \item  Inspired by VIDON \cite{prasthofer2022variableinputdeepoperatornetworks}, which provides the ability to handle irregular data, we extend its functionality to incorporate multiple past timesteps within a defined historical window. To achieve this, we introduce a temporal encoder alongside the spatial encoder, allowing the model to process data from multiple sources, such as probe vehicles and boundary controls, across both spatial and temporal dimensions. Our architecture also incorporates a nonlinear decoder to better capture the complex solution space, enhancing the representational capacity of the standard DeepONet structure.
    
\item We propose a training strategy that promotes robustness to online deployment scenarios by adapting the learning algorithm in a receding horizon fashion that updates the estimate periodically based on new measurements from the probe vehicles.

    \item We provide an integrated uncertainty quantification method that estimates prediction confidence and highlights regions with high aleatoric uncertainty. To this end, we add additional outputs to both the branch and trunk networks of the DeepONet, inspired by \cite{deeponetUQ}.

\end{enumerate}
GitHub link to the implementation of this paper is: \texttt{https://github.com/STC-Lab/ON-Traffic}.

\subsection{Organization of the paper}
The organization of the paper is as follows. After defining the mathematical notations in section II, in Section III, we discuss the methodology, where the mathematical framework of online traffic flow estimation is introduced. In section IV, we discuss the implementation of ON-Traffic. Section V discusses the detailed evaluation of our approach and we end with section VI, the conclusions and future work. 

\section{Notations}
$\mathbb{N}$ denotes the set of all natural numbers. $\mathbb{R}^{+}$ denotes the interval $[0, \infty)$. The symbol $\odot $ stands for element of multiplication of two arrays. The Euclidean norm for a vector $x := \begin{bmatrix}
    x_1 & x_2& \ldots & x_n
\end{bmatrix}\in \mathbb{R}^n$ is denoted by $\mid\mid x\mid\mid_2 = \sqrt{x_1^2 + x_2^2 +\cdots+ x_n^2}$. A uniform probability distribution is denoted $\mathfrak{U}$. If the random variable drawn from the uniform probability distribution admits values between $r_{\mathrm{min}}$ and $r_{\mathrm{max}}$, the probability distribution is denoted by $\mathfrak{U}(r_{\mathrm{min}},r_{\mathrm{max}})$.

\section{ON-Traffic Learning Methodology}
\subsection{Model Input}

Let $x \in \mathbb{X}$ and $t \in \mathbb{T}$ be the variables denoting space and time that take values from compact sets $\mathbb{X}:=[x_{\mathrm{min}}, x_{\mathrm{max}}] \subset \mathbb{R}$ and $\mathbb{T} \subset \mathbb{R}^{+}$. Here, $x_{\mathrm{min}}, x_{\mathrm{max}}$ denote the boundaries of the road. 
We denote the normalized traffic density as $\rho(x,t)$ and normalized velocity as $v(x,t)$.

Let the current time be denoted by $t_c$ where $t_c \in \mathbb{T}$. It is assumed that there always exists a $\Delta_{\mathrm{past}} >0$ and $\Delta_{\mathrm{pred}}>0$ such that $t_c - \Delta_{\mathrm{past}}, t_c + \Delta_{\mathrm{pred}} \in \mathbb{T}$.

\subsubsection{Input from probe vehicles}The probe-vehicles, whose total number at time $t_c$ is denoted by $N(t_c)$, provide local measurements of both normalized traffic density and normalized velocity at discrete positions in the domain $ \mathbb{X}$. These measurements span a past time window $\Delta_\mathrm{past}$ during which the position of a probe vehicle is denoted by $y(t_k)$ with $t_k\in [t_c - \Delta_{\mathrm{past}}, t_c]$. Thus, the probe vehicle dataset, denoted by $u_{\mathrm{probe}}(t_c)$, is defined as follows:





\begin{equation}
\begin{aligned}
    u_{\mathrm{probe}}(t_c) = \Big\{ &\big( \rho(y_i(t_k), t_k), v(y_i(t_k), t_k), y_i(t_k), t_k \big) \;\big|\; \\
    & t_k \in [t_c - \Delta_{\mathrm{past}}, t_c], \; i = 1, \dots, N(t_c), k \in \mathbb{N} \Big\}.
\label{eq:rho_probe_def}
\end{aligned}
\end{equation}

\subsubsection{Boundary inputs}

At the downstream boundary $x_{\mathrm{max}}$, a known boundary condition $\rho_{\mathrm{control}}(x_{\mathrm{max}}, t)$, influenced by a traffic control mechanism, is available. This boundary condition is assumed to be known for both the past horizon $\Delta_{\mathrm{past}}$ and the prediction horizon $\Delta_{\mathrm{pred}}$, and is represented by the set
\begin{equation}
\begin{aligned}
u_{\mathrm{control}}(t_c) = \Big\{ \rho_{\mathrm{control}}(x_{\mathrm{max}}, t_k) \mid \, & \\
t_k \in [t_c - \Delta_{\mathrm{past}}, & \, t_c + \Delta_{\mathrm{pred}}], k \in \mathbb{N} \Big\}.
\end{aligned}
\end{equation}
The combined input for the surrogate model is the combination of the probe measurements and the boundary control and is formulated as,
\begin{equation}
\label{input_eq}
    u_{\mathrm{input}}(t_c) = u_{\mathrm{probe}}(t_c) \cup u_{\mathrm{control}}(t_c).
\end{equation}
\subsection{Formulation of the surrogate operator}
A surrogate model, denoted by $\mathcal{G}_\theta$, is parameterized by the to-be-optimized parameter set $\theta$ (neural network weights and biases), and predicts the normalized traffic density, velocity, and their uncertainties across the spatial domain $[x_{\mathrm{min}}, x_{\mathrm{max}}]$ and the time window $[t_c-\Delta_\mathrm{past}, t_c+\Delta_\mathrm{pred}]$. The input-output relation of the surrogate model 
is defined for all $ (x, t) \in [x_{\mathrm{min}}, x_{\mathrm{max}}] \times 
    [t_c - \Delta_{\mathrm{past}}, t_c + \Delta_{\mathrm{pred}}]$ as



\begin{equation}
\label{surrogate}
\begin{aligned}
\mathbf{y}_{\mathrm{est}}(x, t) 
    &= \mathcal{G}_\theta \big( u_{\mathrm{input}}(t_c) \big)(x, t). 
\end{aligned}
\end{equation}
The model takes the finite set $u_{\mathrm{input}}(t_c)$, defined according to \eqref{input_eq}, as input and produces predicted output $\mathbf{y}_{\mathrm{est}}(x, t) :=\begin{bmatrix}
 \rho_{\mathrm{est}}(x, t) &  v_{\mathrm{est}}(x, t)   
\end{bmatrix}^{\top}$ consisting of the estimated normalized traffic density $\rho_{\mathrm{est}}$ and the normalized velocity field $\rho_{\mathrm{est}}$ for all $ (x, t) \in [x_{\mathrm{min}}, x_{\mathrm{max}}] \times 
    [t_c - \Delta_{\mathrm{past}}, t_c + \Delta_{\mathrm{pred}}]$.

Taking into account the variability of the input data, the output $\mathbf{y}_{\mathrm{est}}(x, t)$ is considered to be drawn from a probability distribution with mean $\hat{\mathbf{y}}(x,t):=\begin{bmatrix}
 \hat{\rho}(x, t) &  \hat{v}(x, t)   
\end{bmatrix}^{\top}$ and variance $\hat{\boldsymbol{\sigma}}(x,t):=\begin{bmatrix} 
         \hat{\sigma}_{\rho}(x, t) & 
        \hat{\sigma}_{v}(x, t) 
    \end{bmatrix}^{\top}$ associated to the normalized traffic density and velocity estimates. 

By determining the appropriate model class of $\mathcal{G}_{\theta}$ and the parameter set $\theta$, one can establish the surrogate operator and use \eqref{surrogate} to yield the mean $\hat{\mathbf{y}}(x,t)$ and the variance $\hat{\boldsymbol{\sigma}}(x,t)$ of the predicted traffic density and velocity profile at any point $ (x, t) \in [x_{\mathrm{min}}, x_{\mathrm{max}}] \times 
    [t_c - \Delta_{\mathrm{past}}, t_c + \Delta_{\mathrm{pred}}]$.

\subsection{Data-Driven and Uncertainty-Quantified Learning}
\subsubsection{Model architecture}\label{sec_arch}
\begin{figure*}[t]
    \centering
    \includegraphics[width=0.8\textwidth, trim=0cm 1cm 0cm 0cm, clip]{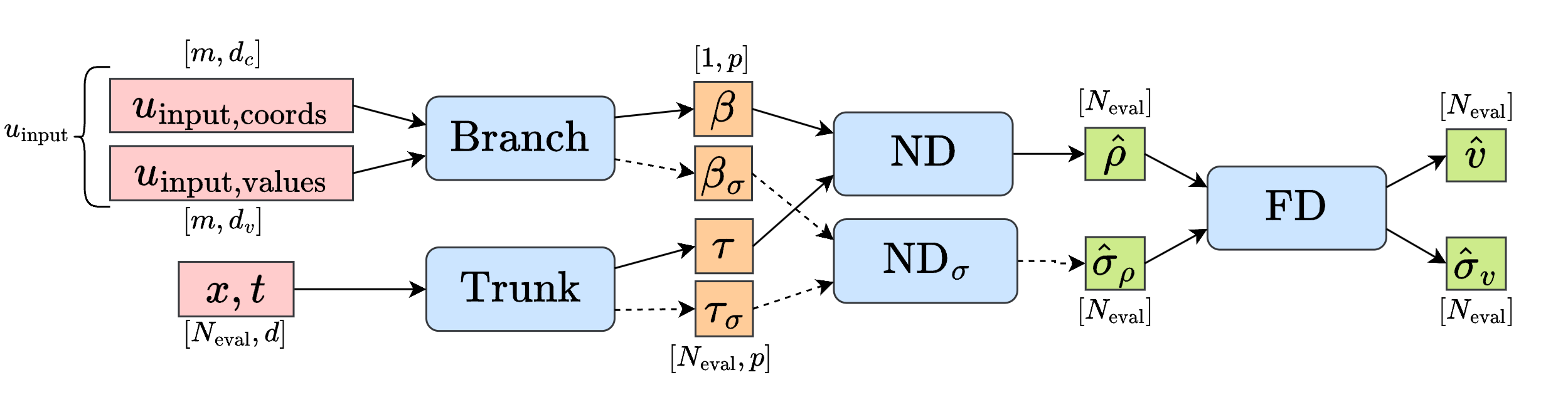}
    \caption{Overview of the proposed model architecture of ON-Traffic.}
    \label{fig:architecture}
\end{figure*}
\begin{figure*}[t]
    \centering
    
    \centering
    \includegraphics[width=0.7\textwidth, trim={0cm 0 0cm 0}, clip]{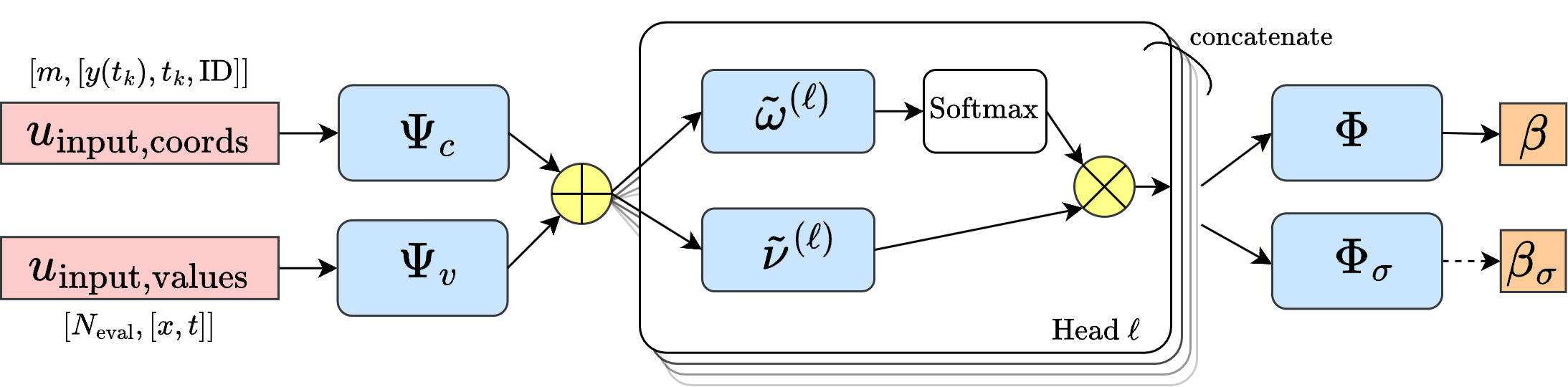}
    \caption{A more detailed illustration of the branch architecture of ON-Traffic based on VIDON \cite{prasthofer2022variableinputdeepoperatornetworks} and extend with a temporal encoder. Additionally we employ separate outputs for the mean predictions and their standard deviation following the approach of \cite{deeponetUQ}.}
    \label{fig:arch_zoomedin}

\end{figure*}
Fig. \ref{fig:architecture} illustrates the proposed architecture of the ON-Traffic model class that consists of a branch network and a trunk network. The trunk network is a Multi-Layer Perceptron (MLP)\footnote{A brief description of MLP and DeepONet is given in Appendix \ref{App_A}.} that generates $p$ basis elements for the output function, consistent with standard DeepONet. The branch network, inspired by the VIDON architecture \cite{prasthofer2022variableinputdeepoperatornetworks}, is modified to handle trajectory data by including time as an additional input. 

The branch network, shown in detail in Fig. \ref{fig:arch_zoomedin}, processes $u_{\mathrm{input}}$, which consists of $m$ observations. Each observation is divided into two components, coordinates $u_{\mathrm{input, coords}}$, a $d_c = 3$ dimensional vector representing location, time, and an identifier (ID) used to distinguish probe vehicles and control inputs, and values $u_{\mathrm{input, values}}$, a $d_v = 2$ dimensional vector capturing the measured normalized density and velocity. 

To encode each observation, the branch network separately applies a coordinate encoder $\Psi_c$ to $u_{\mathrm{input, coords}}$ and a value encoder $\Psi_v$ to $u_{\mathrm{input, values}}$. These encoders are trainable MLPs, and their outputs are summed to obtain a unified representation, \begin{equation} \psi_j = \Psi_c(u_{\mathrm{input, coords}, j}) + \Psi_v(u_{\mathrm{input, values}, j}), \end{equation} where $j \in {1, \dots, m}$ indexes the observations.

The encoded inputs $\psi_j$ are processed through multiple ``heads," which assign attention weights to emphasize relevant observations. For a single head $\ell$, the weight function $\omega^{(\ell)}$ is defined as, \begin{equation} \omega^{(\ell)}(\psi_j) = \frac{\exp\left(\tilde{\omega}^{(\ell)}(\psi_j) / \sqrt{d_{\mathrm{enc}}}\right)}{\sum_{k=1}^m \exp\left(\tilde{\omega}^{(\ell)}(\psi_k) / \sqrt{d_{\mathrm{enc}}}\right)}, \end{equation} where $\tilde{\omega}^{(\ell)}(\psi_j)$ is the output of a head-specific MLP applied to $\psi_j$.

Using these weights, the value representation for head $\ell$ is computed as, \begin{equation} \nu^{(\ell)} = \sum_{j=1}^m \omega^{(\ell)}(\psi_j) \tilde{\nu}^{(\ell)}_j, \end{equation} where $\tilde{\nu}^{(\ell)}_j$ is the output of another head-specific MLP applied to $\psi_j$.

The outputs of all $H$ heads are concatenated into a single vector, \begin{equation} \nu = \begin{bmatrix}
  \nu^{(1)}& \nu^{(2)}& \dots&\nu^{(H)}
\end{bmatrix}^{\top}, \end{equation}
which is passed through the MLPs $\Phi$ and $\Phi_\sigma$, to generate the outputs of the branch network,

\begin{equation}
\begin{aligned}
\beta &= \Phi(\nu), \\
\beta_\sigma &= \Phi_\sigma(\nu),
\end{aligned}
\end{equation}

where $\beta$ and $\beta_\sigma$ are the coefficients that are combined with the trunk network to approximate the density and uncertainty predictions, respectively. This structure ensures invariance to permutations of input observations while accommodating a variable number of input sensors, enabling the operator to be compatible with dynamic input configurations.

Our ON-Traffic model replaces the linear projection of the branch and trunk outputs, with a nonlinear decoder ($\mathrm{ND}$) (c.f., \cite{advancednonlineardecoder, nomad}). This decoder improves performance by addressing the challenges of poor fit and increased complexity associated with a linear projection at the latent space (the space of the outputs from the branch network or trunk network) as its dimension increases. The nonlinear decoder, implemented as an MLP, computes the final outputs using the element-wise product of branch coefficients and trunk basis elements,

\begin{equation}
\begin{aligned}
    \label{eq: G=ND}
    \hat{\rho} &= \mathrm{ND}(\beta \odot \tau),\\
    \hat{\sigma}_\rho &= \mathrm{ND_\sigma}(\beta_\sigma \odot \tau_\sigma).
    \end{aligned}
\end{equation}

ON-Traffic model first predicts the normalized traffic density's mean value $\hat{\rho}$ and its variance $\hat{\sigma}_\rho$. These predictions are then fed into a separate network, denoted as $\mathrm{FD}$, similar to the approaches in \cite{DBLP:journals/corr/abs-2106-03142} and \cite{barreau2021physicsinformedlearningidentificationstate}, which models the fundamental diagram. This fundamental diagram network maps the predicted normalized density's mean value $\hat{\rho}$ to the corresponding normalized velocity's mean value $\hat{v}$,
\begin{equation}
\hat{v} = \mathrm{FD}(\hat{\rho}).
\end{equation}

The uncertainty in velocity $\hat{\sigma}_v$ is derived using a first-order uncertainty propagation technique. Specifically, since $\mathrm{FD}$ is implemented as an MLP, we can compute the gradient of $\hat{v}$ with respect to $\hat{\rho}$ via automatic differentiation, similar to how PINNs compute partial derivatives. The propagated uncertainty $\hat{\sigma}_v$ is then given by,
\begin{equation}
\hat{\sigma}_v = \left| \frac{\partial \mathrm{FD}(\hat{\rho})}{\partial \hat{\rho}} \right|  \hat{\sigma}_\rho.
\end{equation}
\begin{figure*}[t]
    \centering
    \begin{subfigure}{0.45\textwidth} 
        \centering
        \includegraphics[width=\textwidth, trim={0cm 0cm 0cm 0}, clip]{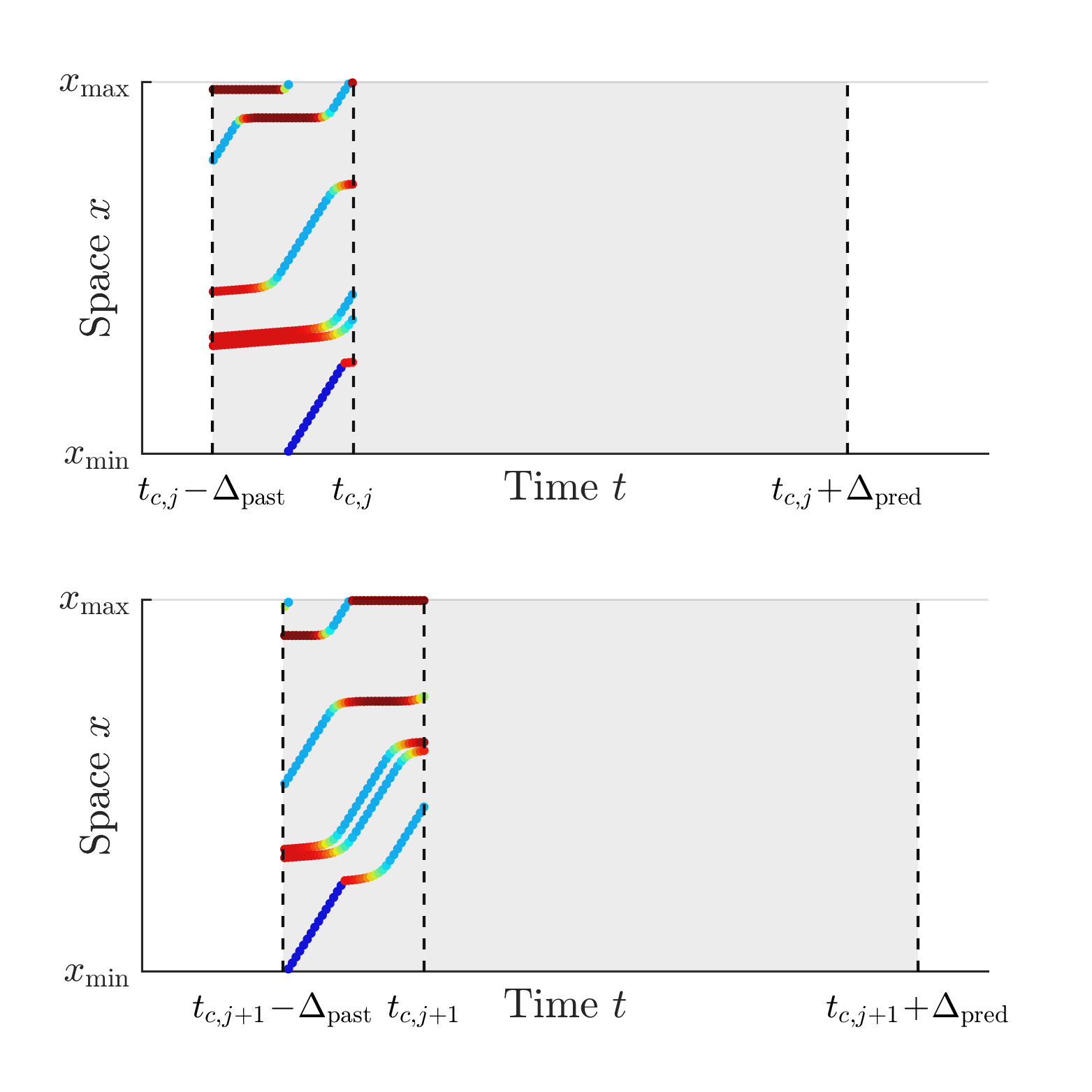}
        \caption{probe inputs}
        \label{fig:probe_inputs_receding2}
    \end{subfigure}
    \hfill 
    \begin{subfigure}{0.45\textwidth} 
        \centering
        \includegraphics[width=\textwidth, trim={0cm 0 0cm 0}, clip]{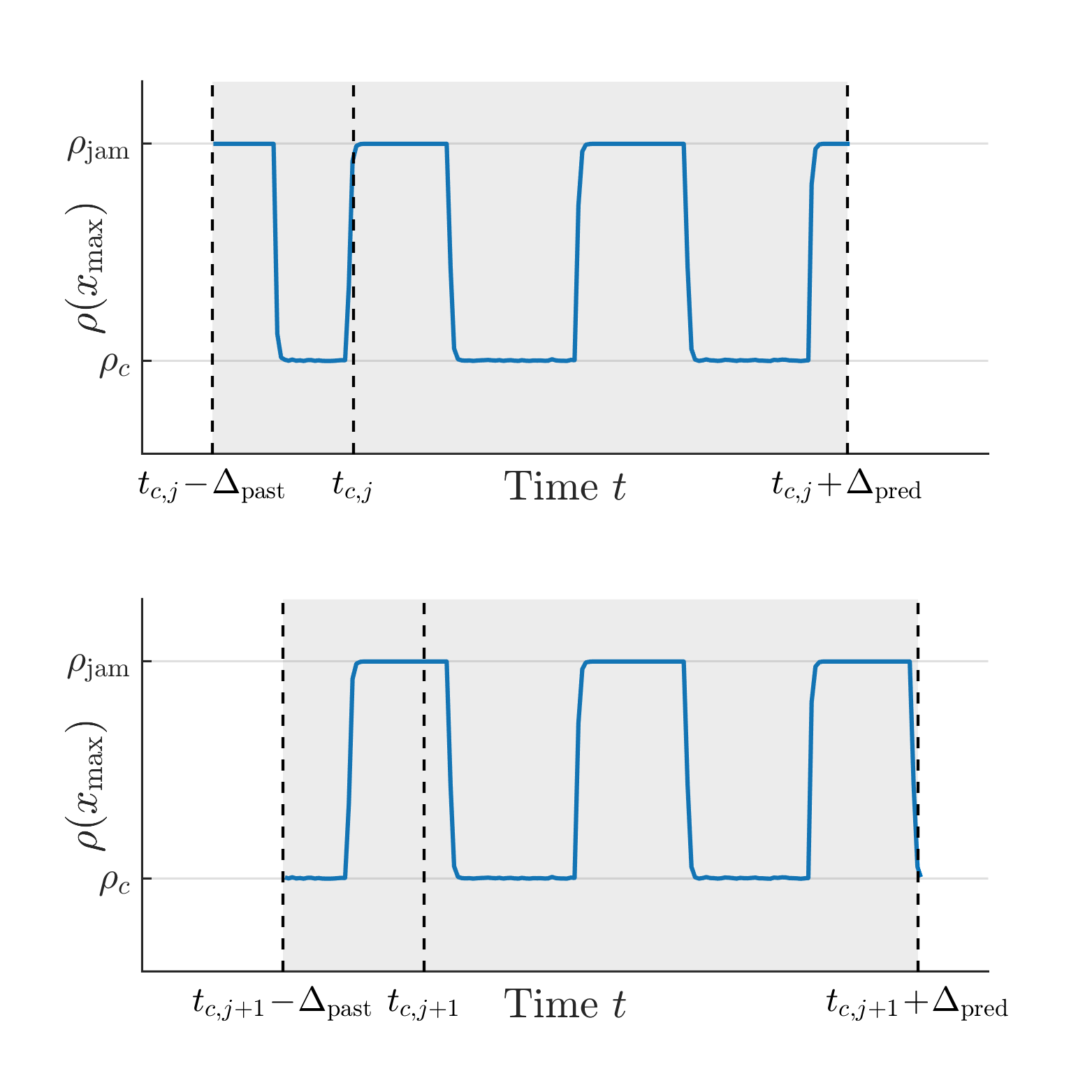}
        \caption{boundary inputs}
        \label{fig:boundary_inputs_receding2}
    \end{subfigure}
    
    \caption{Probe and boundary inputs for the receding horizon implementation of our model. The first row shows the model's input at $t_{c,j}:=t_c+j\Delta_{\mathrm{horizon}}$ and the second row shows the inputs at $t_{c,j+1}:=t_c+(j+1)\Delta_{\mathrm{horizon}}$}
    \label{fig:inputs_receding2}
\end{figure*}

\subsubsection{Optimization problem}
The primary loss function used for regression is the mean squared error, which in operator learning can be defined as,

\begin{equation}
    \mathcal{L}_{\text{data}}^{(n)} = \frac{1}{M_{\text{data}}^{(n)}}\sum_{i=1}^{M_{\text{data}}^{(n)}} 
    \mid \mid \hat{\mathbf{y}}_{i}^{(n)} - \mathbf{y}_{i}^{(n)}\mid \mid_2^2,
\end{equation}
where $\hat{\mathbf{y}}_i^{(n)}$ is the prediction for the $i^{\mathrm{th}}$ data sample of the $n^{\mathrm{th}}$ scenario, and $M_{\text{data}}^{(n)}$ is the number of data points of the $n^{\mathrm{th}}$ scenario. 

To quantify uncertainty, we modify the loss function in a manner similar to that of Gaussian Processes (c.f., \cite{deeponetUQ}). In particular, we use the negative log-likelihood function that assumes prediction errors to be distributed normally around the true value. The resulting loss function is given by

\begin{align}
\mathcal{L}_{\text{data} + \text{UQ}}^{(n)} = &\frac{1}{M_{\text{data}}^{(n)}} \sum_{i=1}^{M_{\text{data}}^{(n)}} \Bigg(
\frac{\mid \mid\hat{\mathbf{y}}_{i}^{(n)} - \mathbf{y}_{i}^{(n)}\mid \mid_2^2}{\mid \mid\hat{\boldsymbol{\sigma}}_{i}^{(n)}\mid \mid_2^2} \nonumber \\
& \quad + \log(2\pi \mid \mid\hat{\boldsymbol{\sigma}}_{i}^{(n)}\mid \mid_2^2)
\Bigg),
\label{eq: data_UQ_loss}
\end{align}

where $\hat{\boldsymbol{\sigma}}_{i}^{(n)}$ is the predicted uncertainty for the  $i^{\mathrm{th}}$ data sample of the  $n^{\mathrm{th}}$ scenario. 

The first term in (\ref{eq: data_UQ_loss}) ensures that the model fits the data, scaled by the predicted uncertainty, meaning that errors are penalized more heavily when the model is confident (i.e., when the uncertainty is small). The second term serves as a regularization component, discouraging the model from predicting large uncertainties everywhere.

The learning problem amounts to determining the parameter set $\theta^{\star}$
that minimizes \eqref{eq: data_UQ_loss} accumulated over $K$ scenarios in total, i.e., 
\begin{equation}
\theta^\star=\arg\min_{\theta}  \sum_{n=1}^{K} \mathcal{L}_{\text{data} + \text{UQ}}^{(n)}.
\label{eq:dual_objective}
\end{equation}

\subsection {Adapted Learning Methodology on a Receding Horizon}
\label{ch: training_strategy}

To enable online deployment, the surrogate model must incorporate possible change in the number of the available probe vehicles, their corresponding velocity and position data to update the estimate of traffic flow. The surrogate model \eqref{surrogate} is suitable for such a task since it implicitly depends on the current time $t_c$. As time progresses, the value of $t_c$ is updated along with the input with new probe measurements and boundary information. Thus, for a given period $\Delta_\text{horizon} >0$, by shifting time-horizon forward by $\Delta_\text{horizon}$, i.e.,
\begin{equation}
    t_c \rightarrow t_c + j \Delta_{\text{horizon}}, j\in \mathbb{N},
\end{equation}
\eqref{surrogate} provides prediction of normalized traffic flow density, normalized velocity field as well as their uncertainty bound on a shifted time-window. See Fig. \ref{fig:inputs_receding2} for an illustration on  how the surrogate model continuously processes new inputs as time progresses. This capability is not inherent to traditional training of operator networks, which are typically trained on initial conditions sampled at the same reference time step (e.g., $t_c = 0$). However, shifting the reference time introduces changes to the characteristics of the solution, causing the learned basis elements to no longer align with the new temporal context. This misalignment leads to significant performance loss unless accounted for. 

ON-Traffic circumvents this challenge by applying a random time shift to the dataset prior to training, as detailed in Algorithm \ref{alg:temporal_shift}. This process of random temporal shift is also visualized in the Appendix \ref{App_C}, Fig. \ref{fig:training_strat_figure}. By modifying the temporal context of the input according to Algorithm \ref{alg:temporal_shift}, the surrogate model is exposed to a variety of shifted scenarios during training, simulating the variability the model will potentially encounter in deployment. This preprocessing step ensures that the operator network learns to generalize across different temporal scales and contexts. As a result, the model maintains robust performance even when the initial condition shifts in time, enabling reliable predictions within a receding horizon framework.

\begin{algorithm}[H]
\caption{Random Temporal Shift for Receding Horizon Training}
\begin{algorithmic}[1]
\Require Dataset $\mathcal{D}$ with $N$ scenarios, each containing:
    \Statex \quad \textbullet \, Branch network input $u_{\text{input}}$:
    \Statex \qquad $\cdot$ Coordinates: $u_{\text{input,coords}}$
    \Statex \qquad $\cdot$ Corresponding values: $u_{\text{input,values}}$
    \Statex \quad \textbullet \, Trunk network input (evaluation grid): $(x,t)$
    \Statex \quad \textbullet \, Ground truth density values: $\rho(x,t)$, 
    \Statex \quad \textbullet \, Parameters: $\Delta_{\text{past}}$, $\Delta_{\text{pred}}$, total time $T$
\Ensure Preprocessed dataset $\mathcal{D}_{\text{shifted}}$
\State Initialize $\mathcal{D}_{\text{shifted}} \gets \emptyset$
\For{each scenario $(u_{\text{input}}, (x,t), \rho(x,t))$ in $\mathcal{D}$}
    \State Sample time $t_c \sim \text{Uniform}([\Delta_{\text{past}}, T - \Delta_{\text{pred}}])$
    \State Extract temporal window for $u_{\text{input}}$:
    \[
    u_{\text{input,coords}} \gets u_{\text{input,coords}}[t_c - \Delta_{\text{past}}, t_c + \Delta_{\text{pred}}]
    \]
    \[
    u_{\text{input,values}} \gets u_{\text{input,values}}[t_c - \Delta_{\text{past}}, t_c + \Delta_{\text{pred}}]
    \]
    \State Extract temporal window for $\rho$:
    \[
    \rho \gets \rho[(x,t) \cap (t_c - \Delta_{\text{past}}, t_c + \Delta_{\text{pred}})]
    \]
    \State Shift temporal coordinates:
    \[
    u_{\text{input,coords}} \gets u_{\text{input,coords}} - (0,t_c,0)
    \]
    \[
    (x,t) \gets (x,t) - (0,t_c)
    \]
    \State Append shifted scenario $(u_{\text{input}}, (x,t), \rho)$ to $\mathcal{D}_{\text{shifted}}$
\EndFor

\Return $\mathcal{D}_{\text{shifted}}$

\end{algorithmic}
\label{alg:temporal_shift}
\end{algorithm}

To ensure our model performs robustly under conditions where the number of probe measurements can vary and is unknown beforehand, we simulate this variability during training by randomly sampling different numbers of probe measurements. This strategy helps the model generalize better to real-world scenarios with unpredictable probe availability. Recent work by \cite{karumuri2024efficienttrainingdeepneural} employs a strategy of randomly sampling evaluation points during training, resulting in each scenario being evaluated at different locations across epochs. Their results show both a reduction in training time and an improvement in generalizability. Our strategy extends this approach by not only randomly sampling the evaluation coordinates passed to the trunk network, but by also randomly sampling the inputs to the branch network. This is not feasible in the standard DeepONet, which inherently assumes the same number and fixed locations for its inputs.

\section{Data Generation}\label{ch: datasets}

\subsection{Godunov Dataset}
\label{ch: numerical_results}
For the numerical data, we consider a highway of length 5 kilometers. We select initial conditions that are relevant to our use case. Since vehicles on highways often travel in queues, we simulate this behavior by generating initial conditions using a multi-step piecewise constant density profiles with uniform randomly sampled parameters. The initial condition is sampled from
\begin{equation}
    \rho(x, 0) = \rho_{\text{init}} + \sum_{k=1}^{N_s} \Delta h_k \cdot \mathbb{I}_{[x_{k-1}, x_k)}(x),
\label{eq: ic}
\end{equation}
where \( \rho_{\text{init}} \) represents the inflow density and is set to \( \rho(x_\mathrm{min}, t) = 0.1 \). The function \( \mathbb{I}_{[x_{k-1}, x_k)}(x) \) is the indicator function, which equals to 1 if \( x \in [x_{k-1}, x_k) \) and 0 otherwise.  The random step height $\Delta h_k$ is sampled from the range
\begin{equation}
\Delta h_k \sim \mathfrak{U}(h_{\text{min}}, h_{\text{max}}),
\end{equation}
where
\begin{equation}
h_{\text{min}} = -\rho_{\text{current}} \text{,} \quad h_{\text{max}} = 1 - \rho_{\text{current}}
\end{equation}
ensure that $\rho(x, 0)$ remains within the bounds $0 \leq \rho(x, 0) \leq 1$. The positions $x_k$ of the step transitions are defined by
\begin{equation}
x_k \sim \mathfrak{U}(x_{k-1} + s_{w, \text{min}}, x_{k-1} + s_{w, \text{max}}),
\end{equation}
where $s_{w, \text{min}}$ and $s_{w, \text{max}}$ specify the minimum and maximum allowed step widths.

\begin{figure}[H]
    \centering
    \begin{subfigure}[t]{0.48\textwidth}
        \centering
        \includegraphics[width=\textwidth, trim={1.5cm 0 1.5cm 0}, clip]{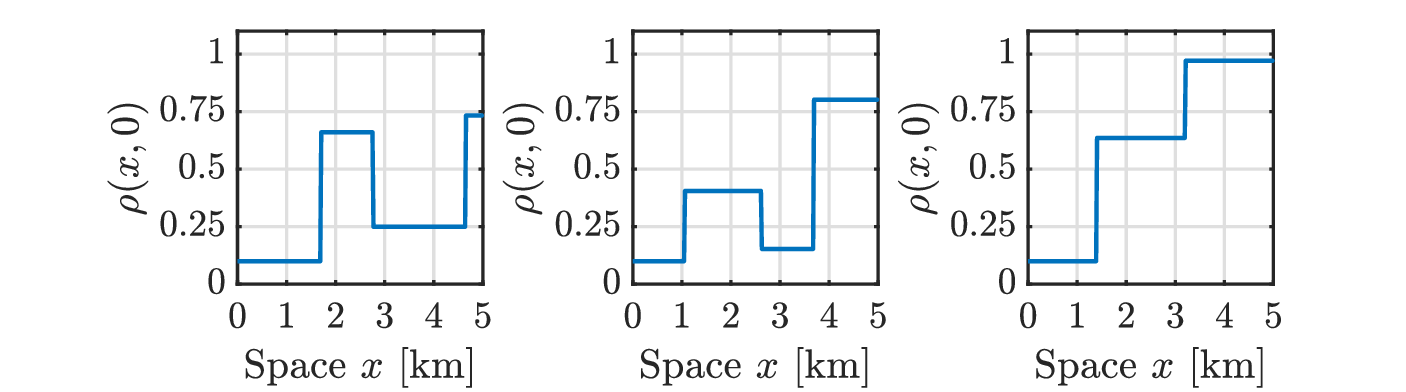}
        \caption{Initial condition realizations.}
        \label{fig:subfig1}
    \end{subfigure}
    \hfill
    \begin{subfigure}[t]{0.48\textwidth}
        \centering
        \includegraphics[width=\textwidth, trim={1.5cm 0 1.5cm 0}, clip]{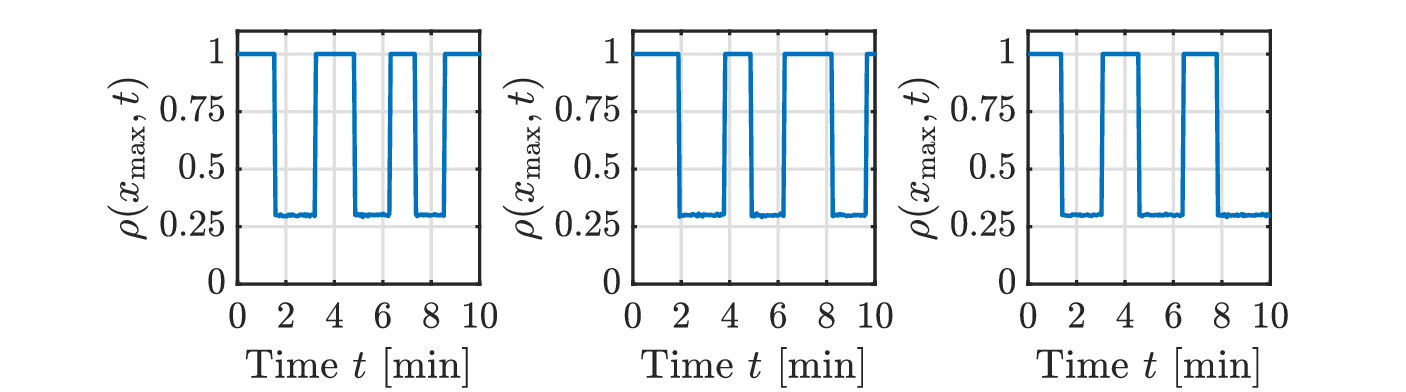}
        \caption{Boundary condition realizations.}
        \label{fig:subfig2}
    \end{subfigure}
    
    \caption{Three realizations from the randomly sampled initial and boundary condition based on (\ref{eq: ic}) and (\ref{eq:control eq}).}
    \label{fig:samples_of_ic_bc}
\end{figure}

For the boundary density at $x_{\mathrm{max}}$, we assume piecewise constant densities resulting from the alternating red and green cycles of a traffic light. Thus the control input $u_{\text{control}}$ is defined as follows
\begin{equation}
u_{\text{control}}(t) =
\begin{cases}
\rho_c, & \text{for } t \in \Delta_{\text{red}, i} \\
\rho_{\text{jam}}, & \text{for } t \in \Delta_{\text{green}, i}
\end{cases}
\label{eq:control eq}
\end{equation}
where $\rho_c$ and $\rho_{\text{jam}}$ are the controlled density values during the red and green cycles, respectively. Each duration $\Delta_{\text{red}, i}$ and $\Delta_{\text{green}, i}$ is independently sampled from a uniform distribution,
\begin{equation}
\begin{aligned}
\Delta_{\mathrm{red}, i} &\sim \mathfrak{U}(\Delta_{\mathrm{min}}, \Delta_{\mathrm{max}}), \\
\Delta_{\mathrm{green}, i} &\sim \mathfrak{U}(\Delta_{\mathrm{min}}, \Delta_{\mathrm{max}}),
\label{eq:duration_sampled}
\end{aligned}
\end{equation}  
where $\Delta_{\mathrm{min}}=1\text{min}$, and $\Delta_{\mathrm{max}}=2\text{min}$ specify the minimum and maximum durations of the red and green light cycles. This sampling introduces variability in the traffic light phases to reflect realistic traffic control behavior. Fig. \ref{fig:samples_of_ic_bc} shows three realizations of the initial conditions and boundary conditions.

In our baseline experiment, we assume an average vehicle density $\bar{\rho} = 40  \text{vehicle/km}$ with a probability of a vehicle being a probe vehicle is $0.03$. The experiment considers a history window of $\Delta_\mathrm{past} = 2 \text{min}$ and a prediction horizon of $\Delta_\mathrm{pred} = 8 \text{min}$. 

\subsection{SUMO Dataset} 
Unlike the numerical dataset we generate, the SUMO simulator is a microscopic traffic simulation tool that models individual vehicle behavior instead of explicitly using the first principle model such as the LWR model. SUMO employs the Intelligent Driver Model (IDM), a second-order car-following model designed to simulate realistic acceleration and deceleration dynamics. The IDM accounts for factors such as desired speed, safe following distance, and the relative velocity between vehicles.

The acceleration $a_i$ of a vehicle $i$ in IDM is given by, 
\begin{equation} a_i = a \left[ 1 - \left( \frac{v_i}{v_0} \right)^4 - \left( \frac{s^\star(v_i, \Delta v_i)}{s_i} \right)^2 \right] \end{equation} 
where $a$ is the maximum acceleration, $v_i$ is the current speed of the vehicle, $v_0$ is the desired speed, $s_i$ is the actual gap to the vehicle ahead, and $\Delta v_i$ is the relative velocity. The term $s^\star(v_i, \Delta v_i)$ represents the desired minimum gap to the vehicle in front, which depends on the current speed $v_i$, and the relative velocity $\Delta v_i$.

In our SUMO dataset, we simulate traffic on a 1 km road segment with a steady inflow of vehicles. At the end of this road, a traffic light regulates vehicle flow, with the green and red durations sampled from a uniform distribution as specified by (\ref{eq:duration_sampled}), with $\Delta_{\mathrm{min}} = 1 \text{ min}$ and $\Delta_{\mathrm{max}} = 2 \text{ min}$. We use the same time windows as for the numerical data, namely $\Delta_\mathrm{past} = 2 \text{ min}$ and $\Delta_\mathrm{pred} = 8 \text{ min}$.

\begin{figure*}[t]
\centering
  \includegraphics[width=\textwidth]{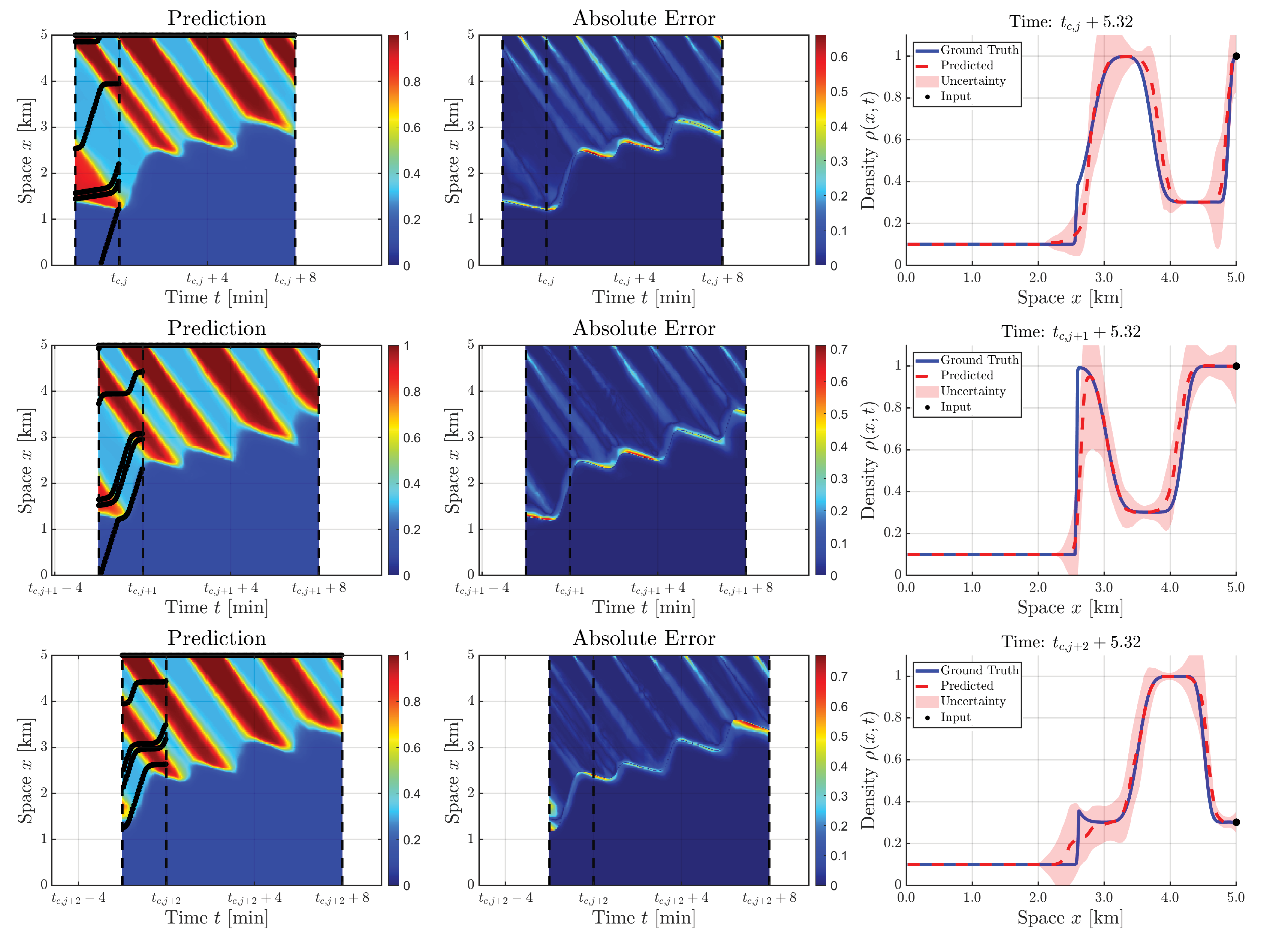}
    \caption{Visualization of one test scenario of the Godunov dataset subject to a receding horizon evaluation. The first column shows ON-Traffic's predictions with the coordinates of the inputs in black, the second column shows the absolute error, and the third column shows the performance of a snapshot in the future. For $t_{c, j+2}$, the predicted uncertainty's average value is $0.0424$ and its peak value $0.2961$. The peak value occurs at a space position of $3.04$ km and a time of $t_c + 4.75$.}
    \label{fig:receding_horizon_3x3_godunov}   
\end{figure*}
\begin{figure*}[t]
\centering 
   \includegraphics[width=\textwidth, trim={4cm 0 4cm 0}, clip]{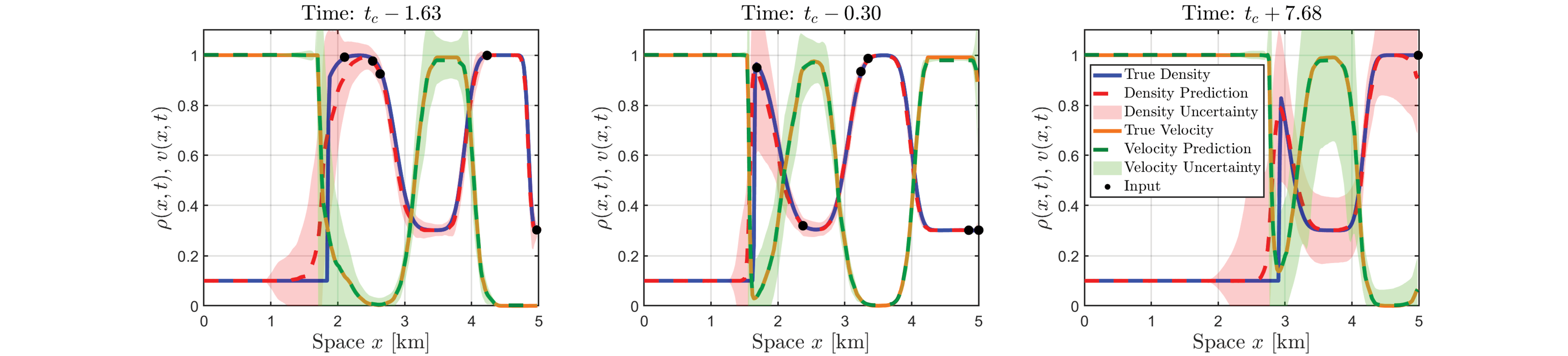}
    \caption{ON-Traffic's results at three selected time instances of the Godunov test set. The blue and orange line represent the true normalized density and normalized velocity values, while the dashed red and green lines show the model's predictions for density and velocity respectively. The shaded regions denotes the 95th percentile of the predicted uncertainty.}
    \label{fig:baseline_time_goduno_velocity}  
\end{figure*}


\section{Evaluation of ON-Traffic}

To evaluate the effectiveness of our approach, we train the proposed ON-Traffic model, as described in section \ref{sec_arch}, by using the datasets introduced in section \ref{ch: datasets} and solving \eqref{eq:dual_objective}. The training is performed in \texttt{PyTorch} using the \texttt{Adam} optimizer \cite{kingma2017adammethodstochasticoptimization} with a batch size of 32 and a learning rate of 0.001. The dataset is generated in 16-bit precision, while the model operates in 32-bit floating point. Also, we split the dataset into 80\% for training and 20\% for validation, while testing is conducted on a separate set of 10,000 scenarios. To improve computational efficiency and reduce memory usage, we employ mixed-precision training. This allows for faster training and optimized GPU performance. On average, the trained model achieves an inference time of approximately 5 milliseconds on an RTX 3080 GPU for making predictions on previously unseen data, rendering it suitable for deployment in online systems.

The remainder of this section evaluates our model’s performance and robustness to ensure real-world applicability. We begin with general performance metrics, followed by analyses of resilience to noise, sensor dropout, and the effect of historical data length. Finally, we assess the model's consistency under time-shifted conditions and validate the reliability of its uncertainty estimates.





\subsection{Evaluating prediction accuracy of the model}

\begin{figure}[h]
    \centering
        \centering
        \includegraphics[width=0.8\linewidth, trim={0cm 0 0cm 0}, clip]{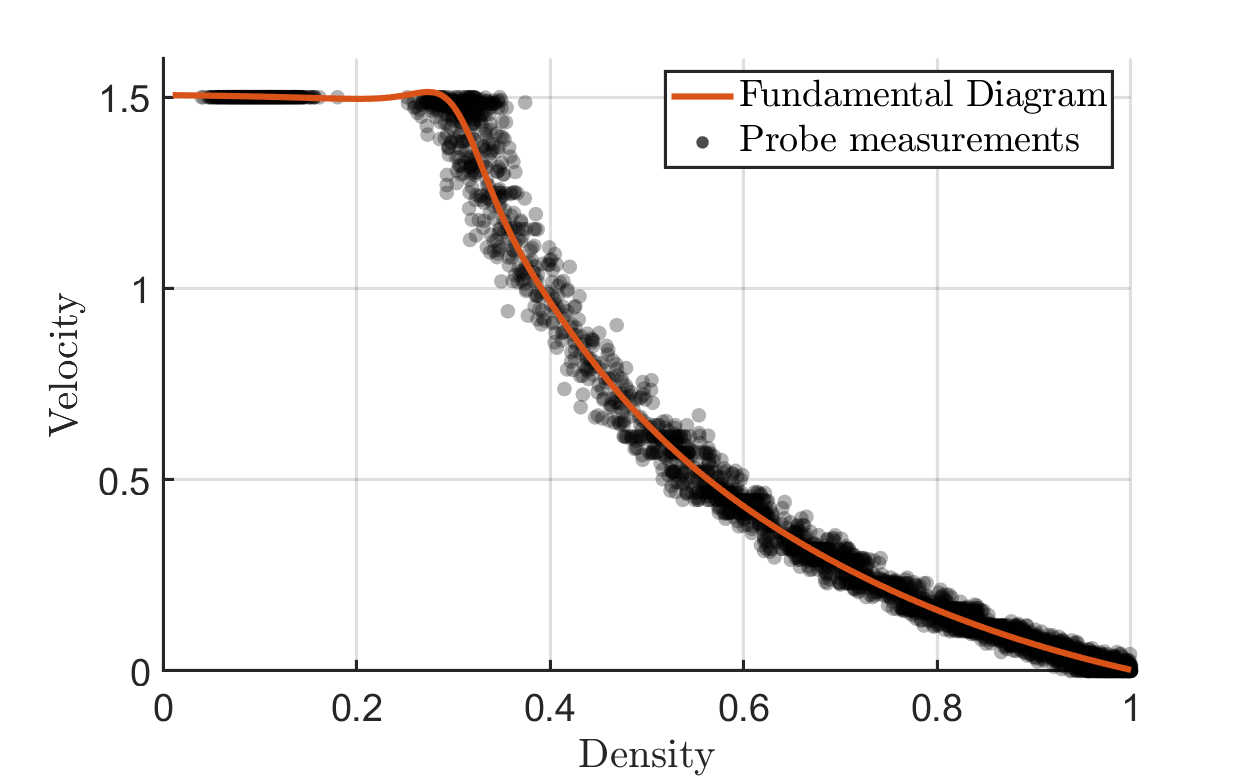}
        \caption{The learned $\mathrm{FD}$ for the Godunov Dataset.}
        \label{fig:FD_godunov}
\end{figure}

\begin{figure}[h]
    \centering
        \centering
        \includegraphics[width=0.8\linewidth, trim={0cm 0 0cm 0}, clip]{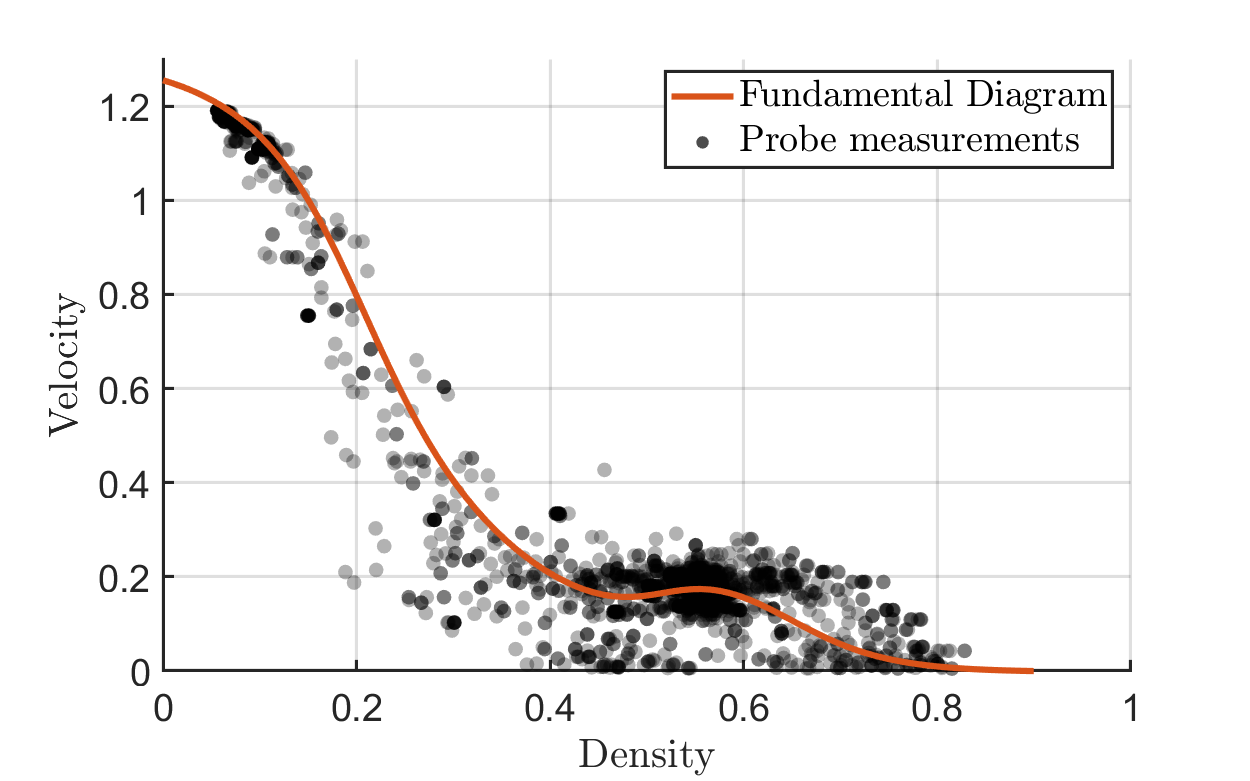}
        \caption{The learned $\mathrm{FD}$ for the SUMO Dataset.}
        \label{fig:FD_sumo}
\end{figure}

\begin{figure}[h]
    \centering
    \includegraphics[width=\textwidth]{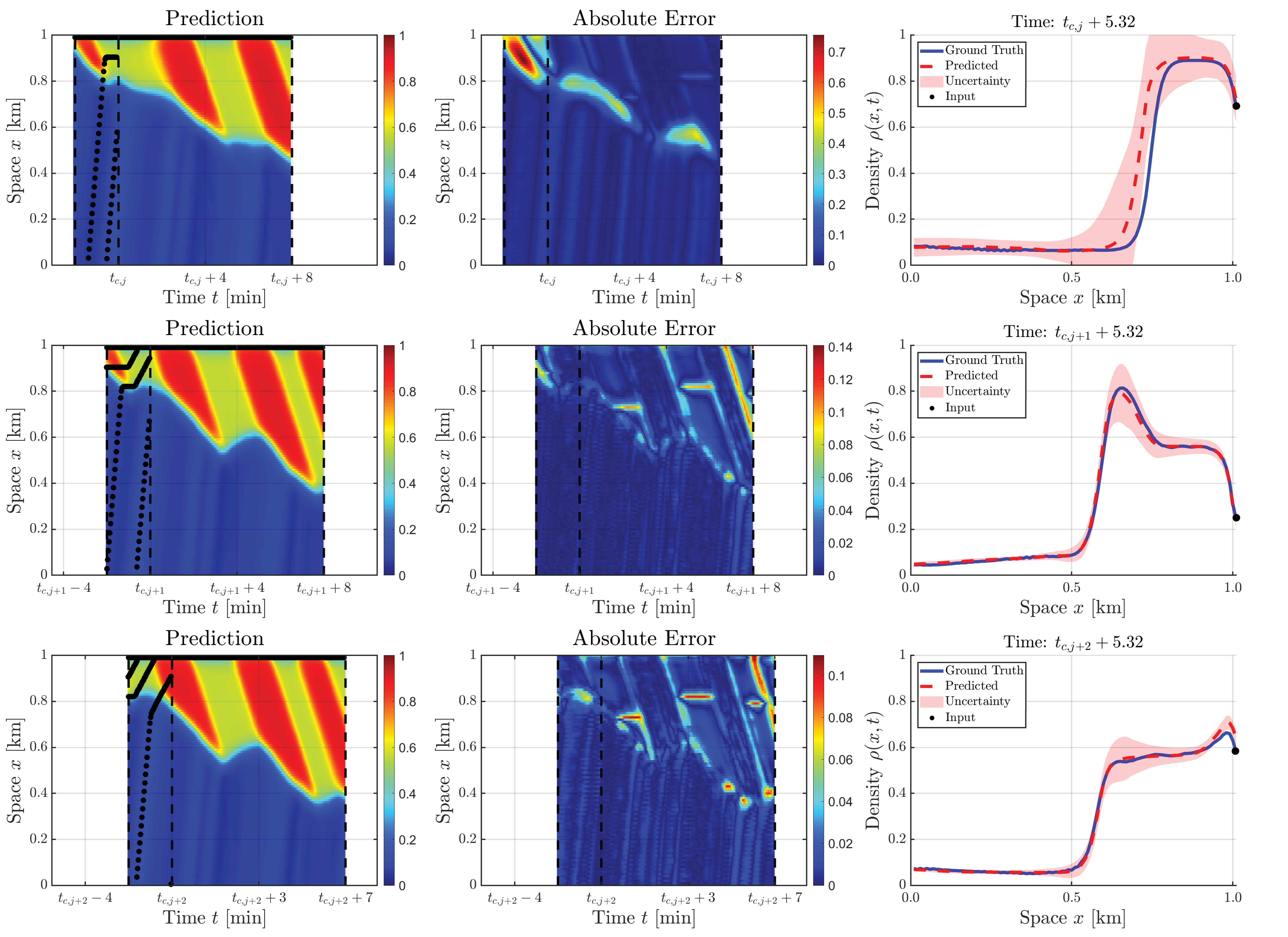}
    \caption{Visualization of one test scenario of the SUMO dataset exposed to a receding horizon evaluation. The first column shows our model's predictions with in black the coordinates of the inputs, the second column shows the absolute error, and the third column shows the performance of a snapshot in the future. For $t_{c, j+2}$, the predicted standard deviation's average is $0.0201$ and its peak value is $0.1343$. The peak occurs at a space position of 0.94 km and a time of $t_c + 0.8$.}
    \label{fig:receding_horizon_3x3_sumo}
\end{figure}

Fig. \ref{fig:receding_horizon_3x3_godunov} and Fig. \ref{fig:baseline_time_goduno_velocity} show the prediction capabilities of our approach for the Godunov dataset. For the SUMO dataset, the results are shown in Fig. \ref{fig:receding_horizon_3x3_sumo}, Appendix \ref{app_b}. The results show that, despite the sparse data, the model achieves a good fit across the datasets. We also visualize the two-standard-deviation confidence bounds in Fig. \ref{fig:receding_horizon_3x3_godunov} and Fig. \ref{fig:baseline_time_goduno_velocity}. These bounds are larger in regions with higher prediction errors, indicating that the model accurately predicts its uncertainty and effectively identifies areas of reduced confidence. Besides the density predictions, also the fundamental diagram $\mathrm{FD}$ is accurately learned and plotted in Fig. \ref{fig:FD_godunov}-\ref{fig:FD_sumo}.

To quantify the model’s accuracy, we evaluated it using mean squared error (MSE) and mean absolute error (MAE) for varying training sample sizes. Tab. \ref{tab:data_type_mse_mae} shows that increasing the training data leads to a significant reduction in error, demonstrating the model’s capacity to learn and generalize effectively with larger datasets. 

\begin{table}[H]
    \centering
    \normalsize 
    \caption{Comparison of MSE and MAE for different data source and training sizes.}
    \renewcommand{\arraystretch}{1.2} 
    \begin{tabular}{l c c c}
        \toprule
        \textbf{Data Source} & \textbf{Training Scenarios} & \textbf{MSE} & \textbf{MAE} \\
        \midrule
        \multirow{3}{*}{Godunov} 
            & 100   & 0.046 & 0.141 \\
            & 1000  & 0.024 & 0.080 \\
            & 10000 & 0.006 & 0.025 \\
        \midrule
        \multirow{3}{*}{SUMO} 
            & 100   & 0.013 & 0.072 \\
            & 1000  & 0.009 & 0.055 \\
            & 10000 & 0.0002 & 0.009 \\
        \bottomrule
    \end{tabular}
    \label{tab:data_type_mse_mae}
\end{table}

\begin{figure}[H]
    \centering
        \centering
        \includegraphics[width=0.8\columnwidth, trim={0.5cm 0 0.5cm 0}, clip]{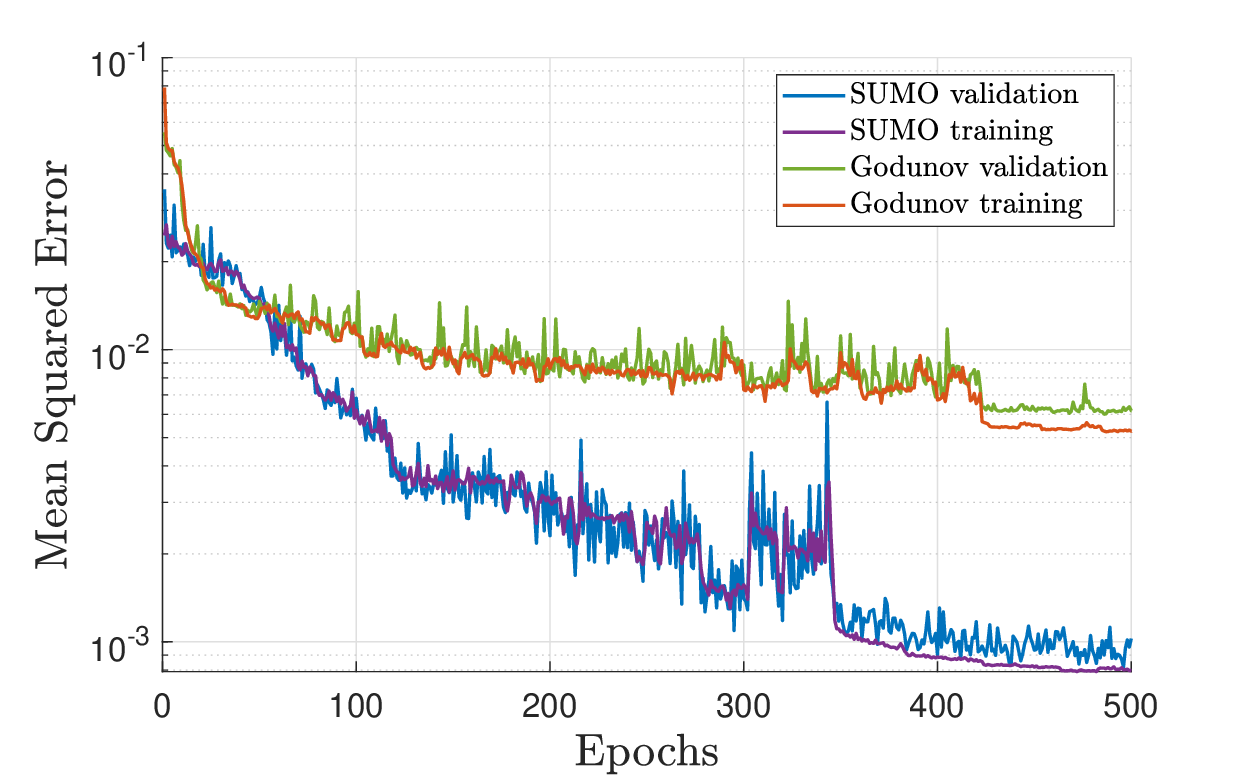}
        
    \caption{The MSE losses of the training and validation set for the Godunov dataset and the SUMO dataset.}
    \label{fig:losses_during_training}
\end{figure}

The training and evaluation loss trajectories, shown in Fig. \ref{fig:losses_during_training}, further highlight the model's learning and generalization behavior. Additionally, we observed that error values are consistently lower for the SUMO dataset, which we attribute to its smoother dynamics compared to the Godunov Dataset. 


\subsection{Analysis of Robustness Against Noise Level and Sensor Dropout Analysis}
\label{sec: noise_and_dropout_analysis}

\begin{figure*}[t]
    \centering
    \begin{subfigure}{\textwidth}
        \centering
        \includegraphics[width=\textwidth]{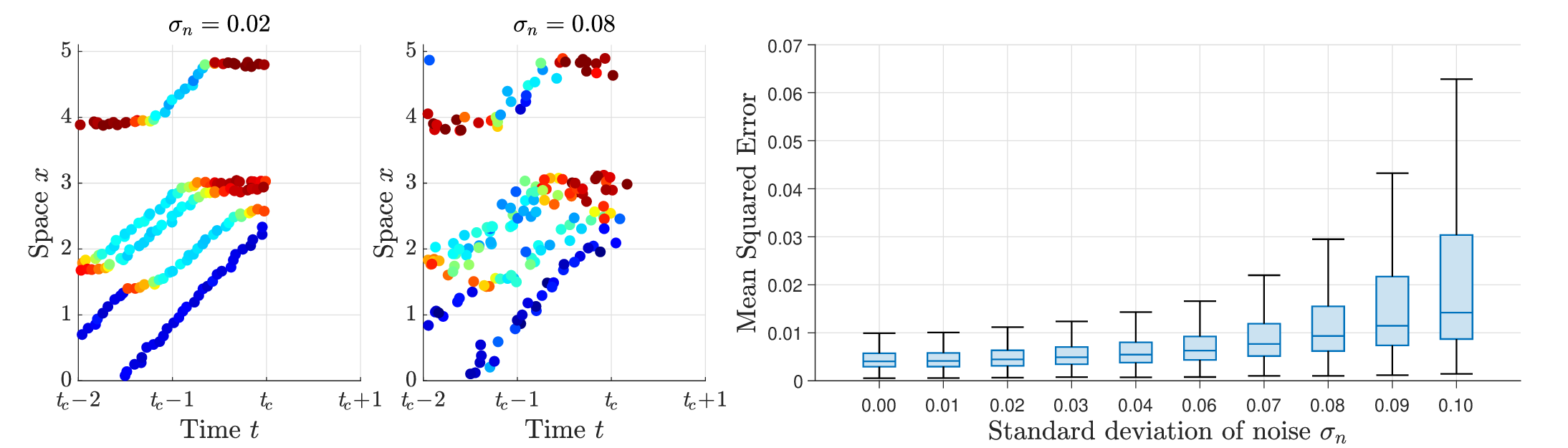}
        \caption{The first two columns show two instances of the probe inputs with added zero mean Gaussian noise with standard deviation $\sigma_n$. The last column shows a boxplot of the mean squared error on the test set for several instances of added noise.}
        \label{fig:noise-analysis}
    \end{subfigure}
    
    \begin{subfigure}{\textwidth}
        \centering
        \includegraphics[width=\textwidth]{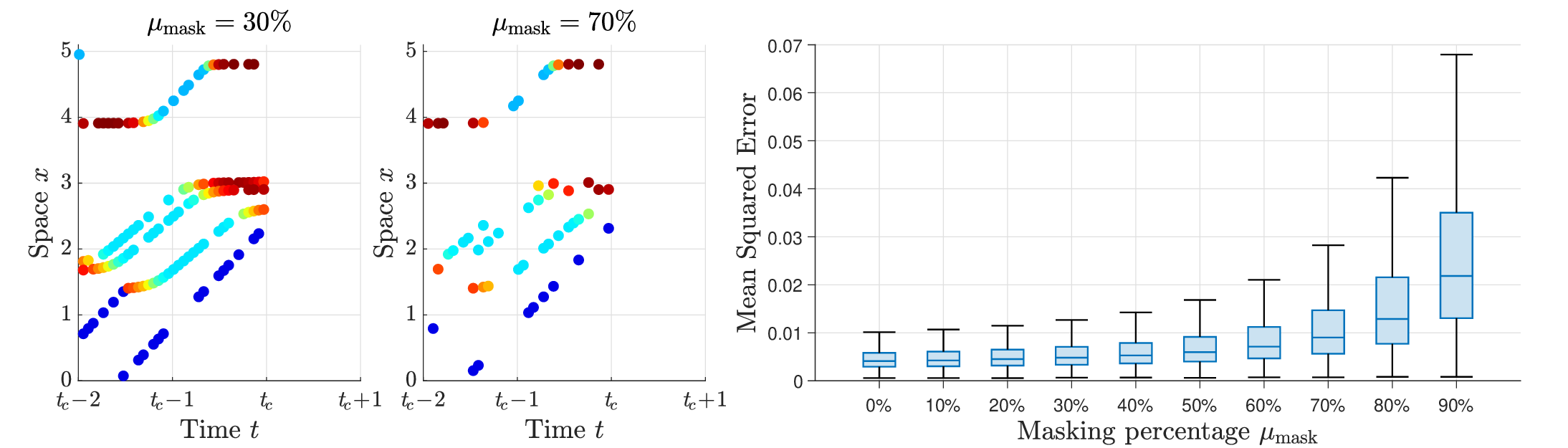}
        \caption{The first two columns show two instances of the probe inputs with a percentage $\mu_{\mathrm{mask}}$ of the probe data being masked to simulate sensor dropout. The last column shows a boxplot of the mean squared error on the test set for several instances of $\mu_{\mathrm{mask}}$.}
        \label{fig:dropout-analysis}
    \end{subfigure}
    
    \begin{subfigure}{\textwidth}
        \centering
        \includegraphics[width=\textwidth]{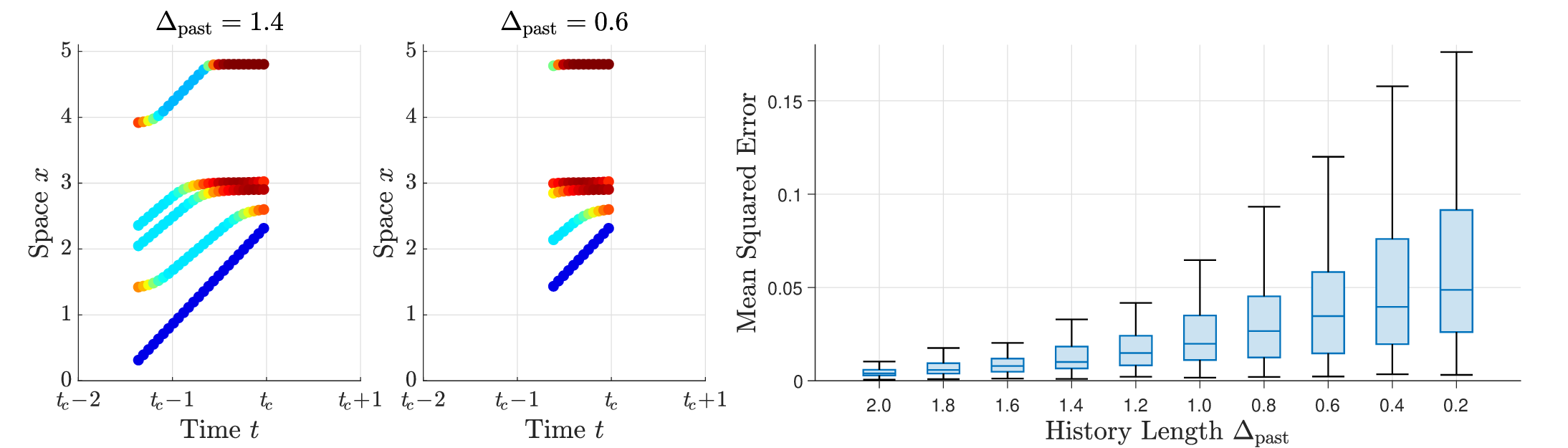}
        \caption{The first two columns show two instances of the probe inputs with certain $\Delta_\mathrm{past}$. The last column shows a boxplot of the mean squared error on the test set for several instances of $\Delta_\mathrm{past}$.}
        \label{fig:timewindow-analysis}
    \end{subfigure}

    \caption{Analysis of noise, dropout and history length effects.}
    \label{fig:combined-analysis}
\end{figure*}

Deploying our model in real-world scenarios involves handling data from a distributed network of vehicle sensors, where various sources of uncertainty can impact the performance. To account for this, we explore two primary sources of uncertainty. First, we introduce zero-mean Gaussian noise with varying standard deviations, $\sigma_n$, to the positional measurements $y_i$ of probe vehicles, as well as to local density measurements. Second, we simulate random sensor dropout, removing a percentage, $\mu_\mathrm{mask}$, of the probe data to model real-world challenges like data loss or sensor failures. 

Fig. \ref{fig:noise-analysis} and \ref{fig:dropout-analysis} illustrate the impact of these uncertainties. The first two columns depict how each scenario affects the input from the probe vehicles, demonstrating the changes introduced by noise and data masking. The final column displays the mean squared error across various scenarios. Notably, our approach demonstrates robustness to these uncertainties up to a certain threshold. In the noise analysis, performance remains stable with minimal error increase for positional noise levels up to a standard deviation of 30 meters. Similarly, in the dropout analysis, the model maintains consistent performance even when up to 30\% of the data is masked. These results indicate that our model can effectively handle moderate levels of noise and data loss, making it well-suited for deployment in real-world environments where sensor reliability and data quality are not guaranteed.

\subsection{Analysis of Accuracy with Varying History Length}
Our hypothesis is that relying only on the sparse initial condition (at one time instant) may not provide enough data to accurately capture the full traffic state. In this section, we investigate the impact of providing the model with additional past time steps and analyze how the length of the historical window, $\Delta_{\mathrm{past}}$, influences the model's ability to reconstruct the density profile. By progressively extending the history of available data, we aim to evaluate how much each additional time step contributes to improving the model's accuracy. Fig. \ref{fig:timewindow-analysis} shows the mean squared error on the validation data as a function of the history length. As anticipated, performance improves with increasing history length. Although the additional data doesn't come from the true initial condition, it proves beneficial because it comes from probe vehicles that measure the past traffic density. This also demonstrates that the temporal encoder of ON-Traffiv effectively captures and interprets the time-encoded information.


\begin{figure}[t]
    \centering
    \includegraphics[width=\columnwidth, trim={0.2cm 0 0.2cm 0}, clip]{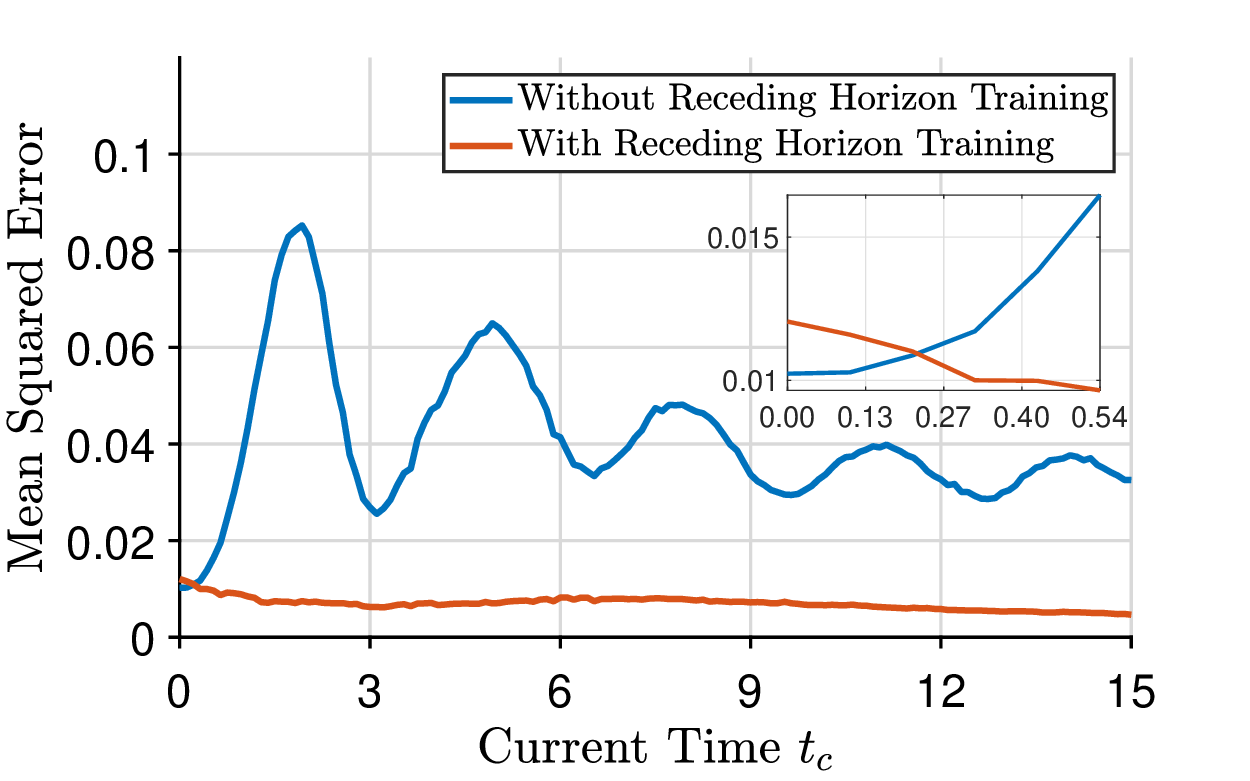}
   \caption{Receding horizon analysis showing the mean squared error over time-shifted versions of the input for models trained with and without our proposed strategy.}
    \label{fig:receding_horizon_result}
\end{figure}

\subsection{Analysis of Accuracy with Receding Horizon Training}
\label{sec: receding_horizon_analysis}

Deploying our model in an online setting requires it to perform consistently well under time-shifted input conditions, which is not inherently guaranteed for standard operator networks. In Fig. \ref{fig:receding_horizon_result}, we compare the performance of ON-Traffic trained with the proposed strategy outlined in Section \ref{ch: training_strategy} against a model trained without this strategy.

The results indicate that the proposed strategy ensures steady performance under time shifts, whereas the baseline model, despite slightly better performance at $t_c = 0$ (the condition on which it was trained), performs poorly and inconsistently under shifted conditions. The oscillations in the baseline model's error may appear because, for certain time shifts, the input becomes closer to the original training condition. This highlights the importance of the proposed training strategy for deployment in a receding horizon fashion.

\subsection{Analysis of Uncertainty Quantification}
A well calibrated uncertainty estimate improves the interpretability and trustworthiness of our model. To validate this, we perform an uncertainty quantification analysis, comparing the model's expected coverage to the observed coverage.

\begin{figure}[t]
    \centering
        \centering
        \includegraphics[width=\columnwidth, trim={0.5cm 0 0.5cm 0}, clip]{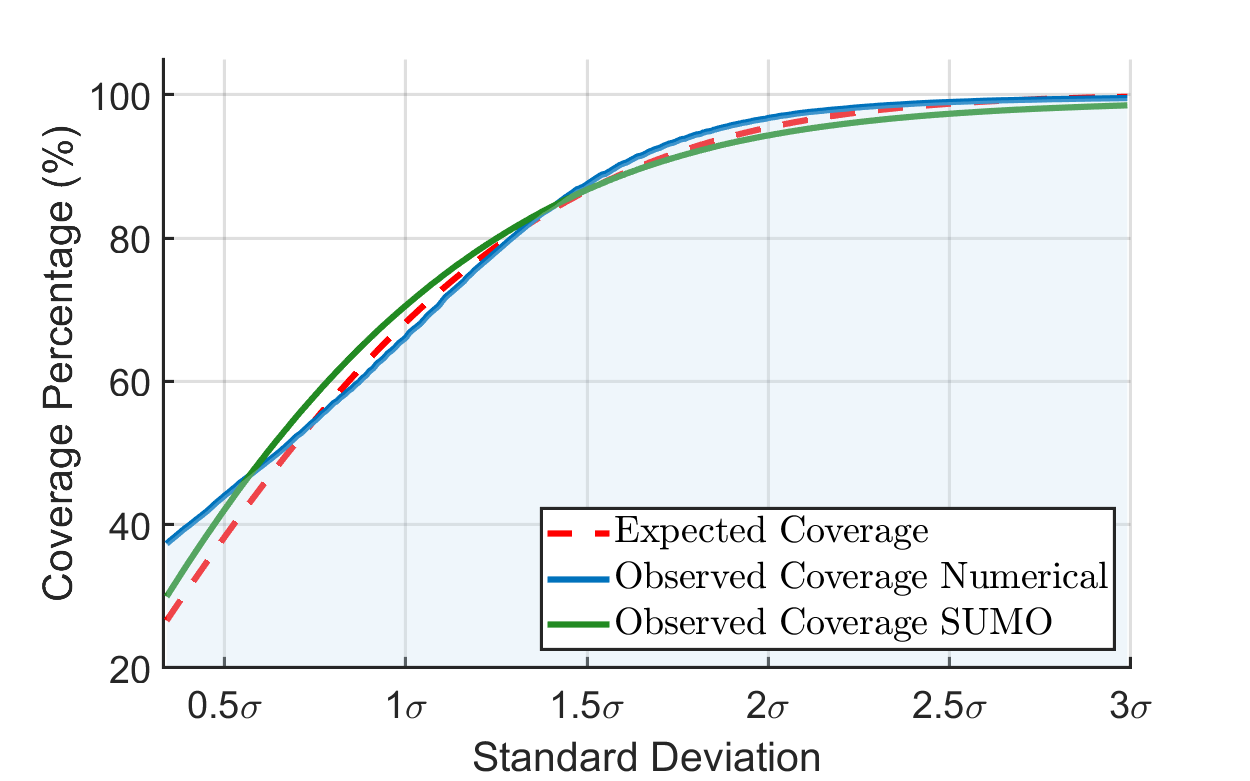}
        
    \caption{Confidence interval coverage plot comparing the model's predicted uncertainty on the numerical data (blue) and the SUMO data (green), with the expected coverage from a normal distribution (dashed red).}
    \label{fig:UQCoverage}
\end{figure}

The expected coverage represents the probability that the true value $\rho$ of the density falls within a range around the predicted value $\hat{\rho}$, given its predictive uncertainty $\hat{\sigma}_\rho$. Since we assume our errors to follow a Gaussian distribution, the expected coverage is expressed as,
\begin{equation} P\left( |\hat{\rho} - \rho| < k\hat{\sigma}_\rho \right) = 2 \times \text{CDF}(k) - 1 \end{equation}
where $\text{CDF}(k)$ is the cumulative distribution function of the standard normal distribution evaluated at $k$ and is illustrated by the dashed red line in Fig. \ref{fig:UQCoverage}.

The observed coverage measures how often the predicted range $\hat{\rho} \pm \hat{\sigma}_\rho$ actually includes the true value $\rho$ over the entire dataset and is expressed as,
\begin{equation} \text{Observed Coverage}(k) = \frac{1}{N} \sum_{i=1}^{N} \mathbb{I}\left(| \rho_i - \hat{\rho}_i | < k\hat{\sigma}_{\rho, i} \right) 
\label{eq:emprical_coverage}
\end{equation}
where $N$ is the number of samples and $\mathbb{I}(\cdot)$ is the indicator function, which is 1 if the condition is true and 0 otherwise.

From Fig. \ref{fig:UQCoverage}, we observe that our model's observed coverage follows well the expected coverage for both the model trained on numerical data and the one trained on SUMO data, indicating that the predicted uncertainty is well-calibrated. Especially for higher values of $\hat{\sigma}_\rho$, the observed coverage aligns closely with the expected coverage, indicating reliable uncertainty estimates in these regions. However, for lower values of $\hat{\sigma}_\rho$, ON-Traffic model tends to overestimate the uncertainty, resulting in observed coverage that exceeds the expected values. This suggests that the model is more conservative at lower uncertainty levels, which can also be visually noted from Fig. \ref{fig:receding_horizon_3x3_godunov}, where in the third column the model predicts rather high uncertainty compared to the actual errors.

\section{Conclusions and Future Work}

This paper presents a new Deep Operator Network based surrogate model designed to tackle the challenges of online traffic flow estimation and prediction using sparse measurements from probe vehicles and control inputs from a downstream traffic light. The framework is capable of processing variable trajectory data, a receding horizon training strategy tailored for online deployment, and a predictive uncertainty quantification method. Through evaluations on both numerical and simulation datasets, we demonstrate that our approach effectively handles the irregularity and variability of probe vehicle data, accommodates time-shifted conditions in a receding horizon manner, and provides well-calibrated uncertainty estimates. These advancements bring the feasibility of deploying operator learning methods for online traffic state estimation in dynamic environments. 

Despite these advancements, we highlight several areas for future work. Currently, our framework is demonstrated on a single road segment with a downstream traffic light. Extending it to more complex and connected road networks with, for example, multiple intersections, merging lanes, and varying traffic control policies would increase its applicability. Additionally, while our model achieves strong performance with simulated data, testing with real-world traffic datasets is essential for validation in practical applications. Real-world data often pose challenges such as poor signal to noise ratio, incompleteness, and high acquisition costs. Integrating domain knowledge through physics-informed learning could bridge the gap between simulated and real-world applications. Finally, our method has shown potential as a building block for predictive control strategies. Future work will explore coupling our framework with control algorithms that can take advantage of the predictive capabilities and uncertainty estimates provided by our model. Through addressing these research directions, ON-Traffic has the potential to be the promising fist step in establishing operator learning methods as a cornerstone for real-time, adaptive traffic management systems.







\newpage

\appendices

\section{Introductory discussions on PINNs and DeepONets}\label{App_A}

The recently proposed PINNs \cite{pinn1}, \cite{pinn2}, belong to a class of neural networks that incorporate the governing physical laws, typically in the form of differential equations, directly into the learning process through the loss function. For a 1-dimensional PDE a PINN would map coordinates $(x,t)$ into the solution $u(x,t)$ via a feed forward neural network, usually a multi-layer perceptron (MLP), where a single layer can be expressed as 
\begin{equation}
    y_i  = \phi{(W_i x_i + b_i)},
\label{eq: mlp}
\end{equation}
where $ \mathbf{x}_i \in \mathbb{R}^{n_{i-1}}$  is the input to layer $i$, $W_i \in \mathbb{R}^{n_i \times n_{i-1}} $ are the weights, $ \mathbf{b}_i \in \mathbb{R}^{n_i} $ are the biases, and $\phi $ is a nonlinear activation function. Via automatic differentiation, the partial derivatives of the solution with respect to its inputs, i.e. space $x$ and time $t$, can be calculated. These partial derivatives are used to calculate the residual of the PDE which is combined with data to form the loss function.

While PINNs directly solve for the solution of a PDE, DeepONets \cite{luludeeponet}, based on the universal approximation theorem for operators \cite{universaloperators}, learn mappings between function spaces. Instead of predicting the solution for a specific condition, DeepONets aim to learn the operator $\mathcal{G}$, that maps one function (e.g., an initial condition) to another (e.g., the solution of the PDE). This makes DeepONets inherently more flexible, as they can generalize to different input functions. DeepONets achieve this by employing two networks with different functions. The $\text{Branch}$ network takes as input the function $u_0$, such as an initial condition, sampled at $m$ discrete points. It encodes these samples into a vector of $p$ coefficients,
\begin{equation}
    \beta = \text{Branch}(u_0(x_1), u_0(x_2), \dots, u_0(x_m)).
\end{equation}
The $\text{Trunk}$ network takes in $N_{\text{eval}}$ coordinates at which to evaluate the output function. These coordinates, denoted as $\xi = \{(x_i, t_i)\}_{i=1}^{N{\text{eval}}}$, represent points in the space-time domain. The Trunk network produces a vector of $p$ basis elements for each input coordinate
\begin{equation}
    \tau = \text{Trunk}(x, t).
\end{equation}
In vanilla DeepONet the solution is then obtained by a linear projection of the branch's coefficients and the trunk's basis functions,
\begin{equation}
    \label{eq: G=b*t}
    \mathcal{G}(u_0)(x,t) = \sum_{i=1}^p \beta_i \tau_i.
\end{equation}
Since the output of the DeepONet depends on the input function, it can generalize to unseen initial conditions unlike PINNs. Additionally, unlike traditional numerical methods, both PINNs and DeepONets are not restricted to a regular grid, allowing them to be evaluated at arbitrary points, not just those used during training.

\onecolumn

\newpage
\section{Prediction results using SUMO dataset}\label{app_b}

\newpage
\section{Processing of Training Data-Random Temporal Shift}\label{App_C}
\vspace{+5ex}
\begin{figure}[h]
    \centering
    \begin{subfigure}{0.9\textwidth}
        \centering
        \includegraphics[width=\textwidth, trim={4cm 0 4cm 0}, clip]{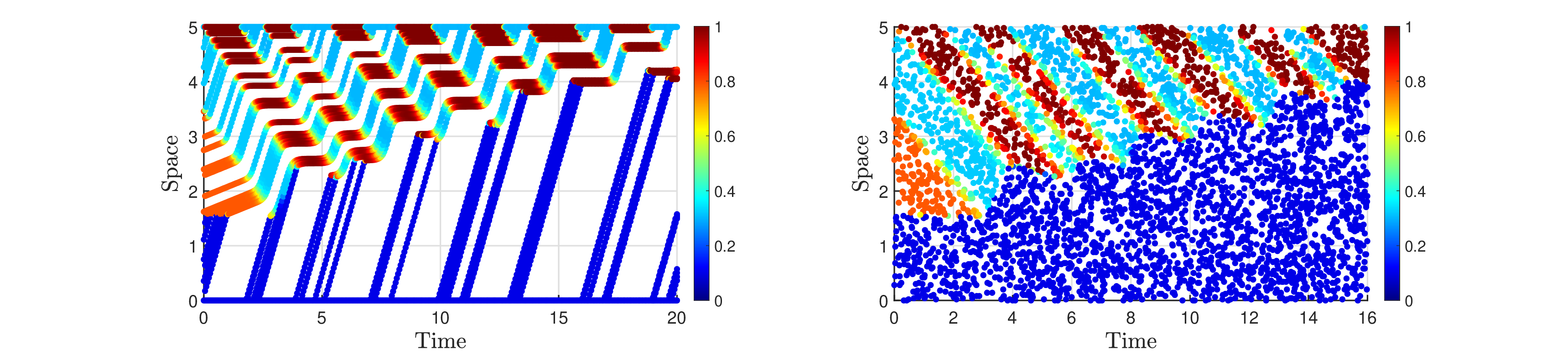}
        \caption{Raw dataset}
        \label{fig:branch_inputs}
    \end{subfigure}

    \begin{subfigure}{0.9\textwidth}
        \centering
        \includegraphics[width=\textwidth, trim={4cm 0 4cm 0}, clip]{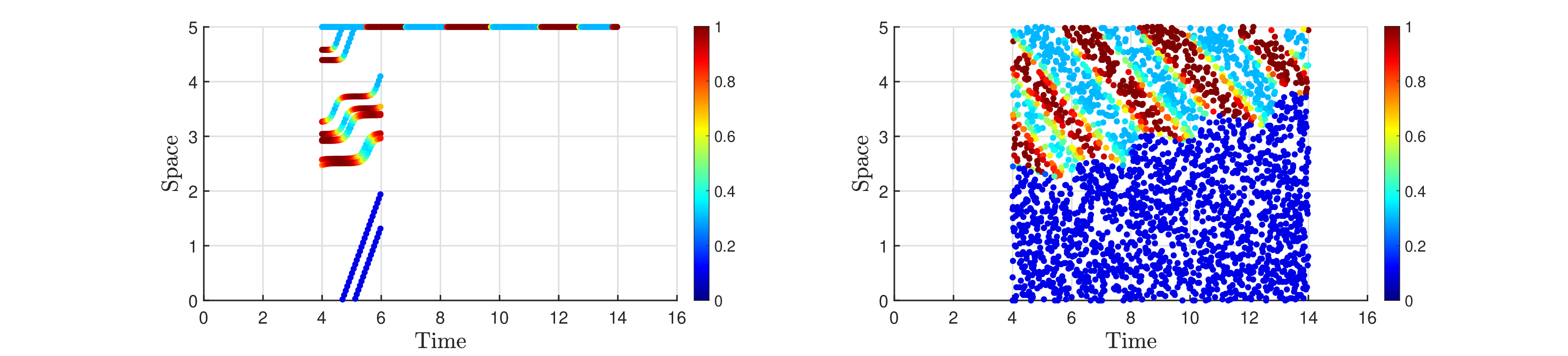}
        \caption{After sampling which vehicles will act as probes, we sample a reference time step $t_c=6$. Then for $u_{\mathrm{input}}$, we keep all probe data between $[t_c-\Delta_\mathrm{past}, t_c]$ and all boundary control data between $[t_c-\Delta_{\mathrm{past}}, t_c+\Delta_\mathrm{pred} ]$. For the target data we keep everything between $[t_c-\Delta_\mathrm{past}, t_c+\Delta_\mathrm{pred}]$.}
        \label{fig:output_sensors}
    \end{subfigure}

    \begin{subfigure}{0.9\textwidth}
        \centering
        \includegraphics[width=\textwidth, trim={4cm 0 4cm 0}, clip]{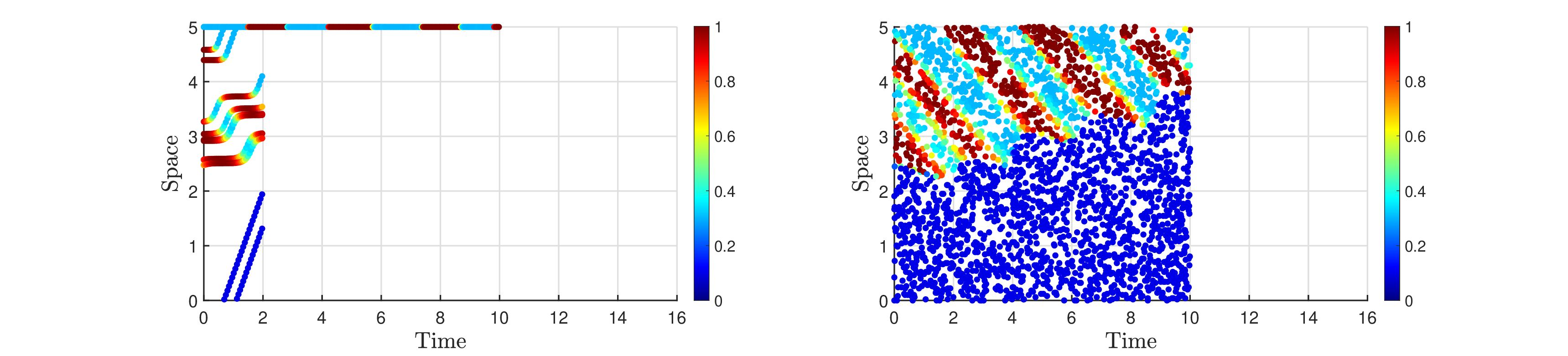}
        \caption{We shift back all data by the selected reference time step $t_c$.}
        \label{fig:filtered_inputs}
    \end{subfigure}

    \begin{subfigure}{0.9\textwidth}
        \centering
        \includegraphics[width=\textwidth, trim={4cm 0 4cm 0}, clip]{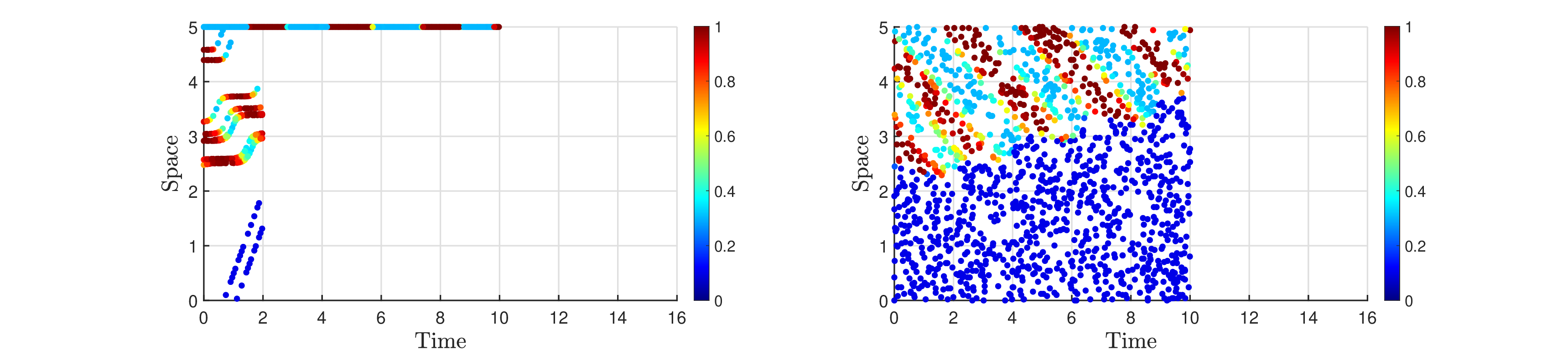}
        \caption{Finally, we randomly select data from both the input as well as our target.}
        \label{fig:filtered_outputs}
    \end{subfigure}

    \caption{Our data processing steps of our training strategy visualized in order. The left column holds the input to the operator and the right column holds the targets used during training.}
    \label{fig:training_strat_figure}
\end{figure}

\end{document}